\documentclass[10pt,twocolumn,letterpaper]{article}

\usepackage[pagenumbers]{cvpr} 

\usepackage{graphicx}
\usepackage{amsmath}
\usepackage{amssymb}
\usepackage{booktabs}
\usepackage{bbding}

\usepackage[pagebackref,breaklinks,colorlinks]{hyperref}

\usepackage[capitalize]{cleveref}
\crefname{section}{Sec.}{Secs.}
\Crefname{section}{Section}{Sections}
\Crefname{table}{Table}{Tables}
\crefname{table}{Tab.}{Tabs.}

\def\cvprPaperID{*****} 
\def\confName{CVPR}
\def\confYear{2023}

\begin{document}

\title{BundleRecon: Ray Bundle-Based 3D Neural Reconstruction
}

\author{Weikun Zhang,\ \ Jianke Zhu\thanks{Corresponding author}\\
Zhejiang University\\
{\tt\small \{zhangwk, jkzhu\}@zju.edu.cn}
}
\maketitle

\begin{abstract}
With the growing popularity of neural rendering, there has been an increasing number of neural implicit multi-view reconstruction methods. While many models have been enhanced in terms of positional encoding, sampling, rendering, and other aspects to improve the reconstruction quality, current methods do not fully leverage the information among neighboring pixels during the reconstruction process. To address this issue, we propose an enhanced model called BundleRecon. In the existing approaches, sampling is performed by a single ray that corresponds to a single pixel. In contrast, our model samples a patch of pixels using a bundle of rays, which incorporates information from neighboring pixels. Furthermore, we design bundle-based constraints to further improve the reconstruction quality. Experimental results demonstrate that BundleRecon is compatible with the existing neural implicit multi-view reconstruction methods and can improve their reconstruction quality.

\end{abstract}


\section{Introduction}
\label{sec:introduction}

Multi-view reconstruction is a fundamental problem in the fields of computer vision. The reconstruction quality of traditional methods is limited by the accuracy of feature matching or the voxel resolution of those explicit representation methods. With the introduction of neural rendering \cite{nerf,sotanr,nerf++,mip-nerf}, many researchers have switched from traditional methods to learning-based neural implicit representation approaches. These methods have an impressive performance on geometry reconstruction, as the implicit function can represent the geometry continuously with relatively low memory cost and high spatial resolution \cite{unisurf,neus}. It is worthy of noting that NeRF~\cite{nerf} was originally intended to solve the problem of novel view synthesis. Meanwhile, its geometry is extracted using an arbitrary level set of the density function, which makes the geometry estimated by NeRF to be less accurate \cite{nerf,volsdf}. To address this issue, researchers replace the density function by other implicit functions such as the occupancy network or the SDF network, from which they can obtain a more precise object representation.

\begin{figure}[t]
    \centering
    \includegraphics[width=7cm]{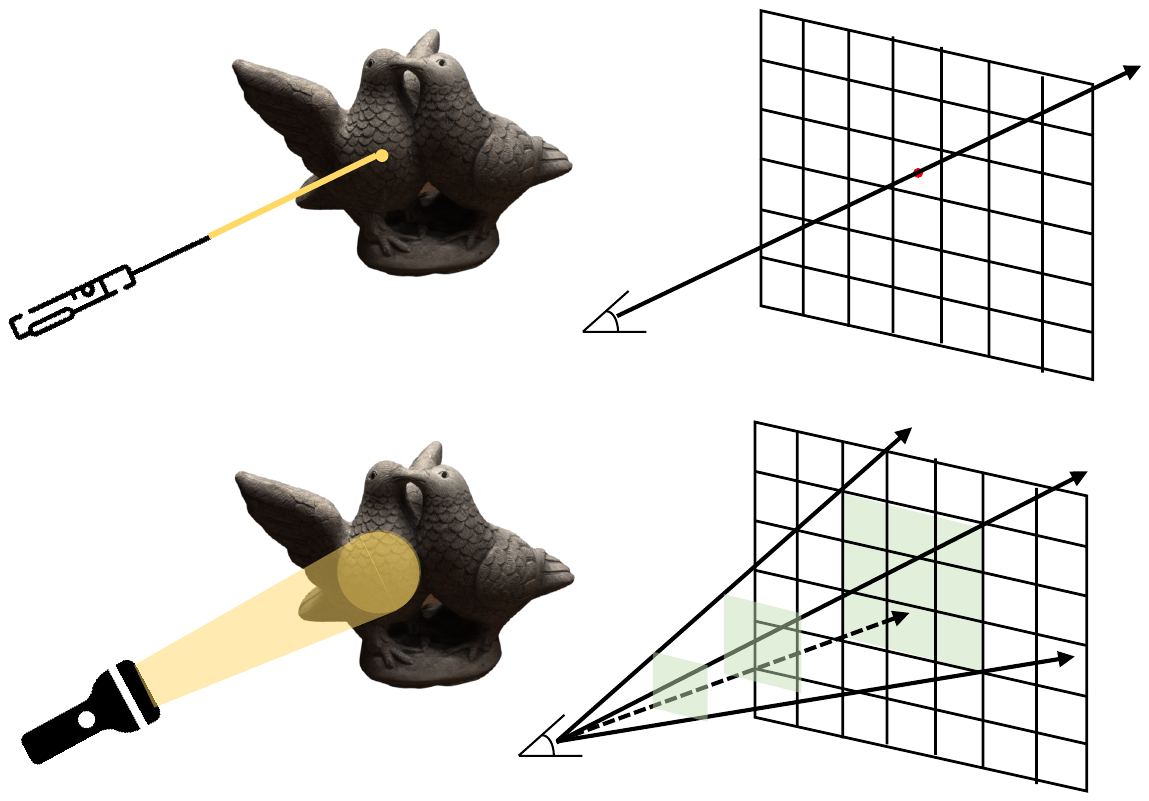}
    \caption{Traditional neural implicit representation methods work like illuminating the object by a laser pen, which provides the limited local information on the illuminated area. In contrast, BundleRecon employs a bundle of rays, which is similar to a flashlight so that the larger illuminated area provides more local information. This is beneficial to the reconstruction process.} 
    \label{fig:diagram}
\end{figure}

For neural implicit representation methods, the sampling process is of vital importance because the selection of sampled points directly affects the quality of the reconstruction. Some studies prefer to the dense sampling near the surface area \cite{volsdf,neurecon-w}, while others adopt a coarse-to-fine sampling process to cover a broader space \cite{nerf,neus,unisurf}. Regardless of the sampling method, they all involve with randomly selecting pixels and using the corresponding ray to sample points. However, this commonly used approach only considers the longitudinal information along the ray while the transversal information between the rays is typically ignored. Therefore, the discrete pixels may result in a loss of information between neighboring pixels.

Inspired by the scenario of using a flashlight to illuminate an object in the dark, we propose BundleRecon to solve the above problem. As shown in Fig. \ref{fig:diagram}, the traditional neural implicit representation methods work like illuminating the object by single ray. Instead, we employ a flashlight, i.e., a bundle of rays. This is because a larger patch of pixels in the field of view provides more useful information. This is more conducive to reconstruction process compared to a single pixel.

Additionally, we have designed bundle-based loss functions to adapt BundleRecon. By utilizing ray bundles to sample pixel patches, we extract bundle-shaped information and feed it to the neural network. Our loss functions make use of statistical information, such as mean and variance, from the bundle-shaped output, while utilizing its convolutional features for supervision. 

We integrate our method into the existing method like IDR\cite{idr} and NeuS\cite{neus} and evaluate them on DTU \cite{dtu} and BlendedMVS \cite{blendedmvs}. The experimental results show that using the information between neighboring pixels can enhance the reconstruction quality.

The contributions of our work are as follows:

\begin{itemize}
\item A ray bundle-based 3D neural reconstruction model is proposed to improve the reconstruction quality by exploiting the information between neighboring pixels.

\item The bundle-based loss functions are designed for BundleRecon to further improve the reconstruction quality.

\item The extensive experiments demonstrate the compatibility and effectiveness of our model, which can be used to improve the reconstruction quality of existing neural implicit multi-view reconstruction methods.
\end{itemize}

\section{Related Work}
Reconstructing objects from multiple images has been intensively studied for over thirty years. We briefly review the related work in the following.
\label{sec:realted}
\subsection{Traditional Multi-view Reconstruction} Multi-view stereo is widely used in reconstruction and can be divided into point cloud reconstruction, volumetric method and depth map approach \cite{mvsnet}. We take the depth map reconstruction as an example to introduce the pipeline of traditional multi-view reconstruction. Firstly, the multi-view input images are processed using structure from motion (SFM) \cite{sfm,sfm2} to generate the camera parameters and sparse 3D points. Secondly, multi-view stereo (MVS) \cite{mvs,mvs2,mvs3} is applied to obtain the depth information. Finally, the resulting depth maps can be fused into point clouds \cite{depth-fusion} that can be used by surface reconstruction methods \cite{poisson-surface-reconstruction}. Although the traditional pipeline has some inherent drawbacks, such as error accumulation through multiple steps, it treats the input image as a whole and considers the information between pixels. In contrast, many neural implicit multi-view reconstruction methods only select a limited number of discrete pixels from the input images, thereby ignoring the information between pixels. Besides, there are some methods that explicitly represent the object. For instance, voxel grids are commonly used to representing objects \cite{voxel1,voxel2} even if the object is of arbitrary topology. However, the reconstruction quality is closely related to the resolution of the grids, which is limited by the system memory.

\subsection{Neural Radiance Fields} 
The proposal of NeRF \cite{nerf} has drawn increasing attention to the implicit representation of scenes. Many researchers have improved NeRF in several ways, such as positional encoding \cite{mip-nerf}, sampling \cite{volsdf,unisurf,neurecon-w,mip-nerf}, and rendering \cite{neus,plenoxels,PlenOctree}. In contrast to methods that simply sample along the ray, Barron \textit{et al.} have proposed a cone-based sampling model, named Mip-NeRF \cite{mip-nerf}, which solves the aliasing problem by casting a cone from each pixel. While the shape of the cone is very similar to our ray bundle, Mip-NeRF's underlying principle is still based on sampling a single ray that corresponds to a single pixel, without considering information from surrounding pixels. In addition, there is work that applies NeRF to dynamic scenes and outdoor scenes \cite{blocknerf,nerf-w,hanerf}.

NeRF was originally proposed to address the problem of novel view synthesis. However, it has since inspired researchers to explore new possibilities in 3D reconstruction. By leveraging the implicit density and color information captured in radiance fields, they discovered that it is possible to reconstruct the object using marching cubes. However, this approach often leads to poor reconstruction quality due to the fact that NeRF's density is obtained from arbitrary level sets. To overcome this limitation, researchers have begun to improve neural radiance fields by using SDF or other implicit representations to extract the geometric information of the scene, in hopes of achieving more precise 3D reconstruction results. 

\begin{figure*}[htbp]
    \centering
    \includegraphics[width=16cm]{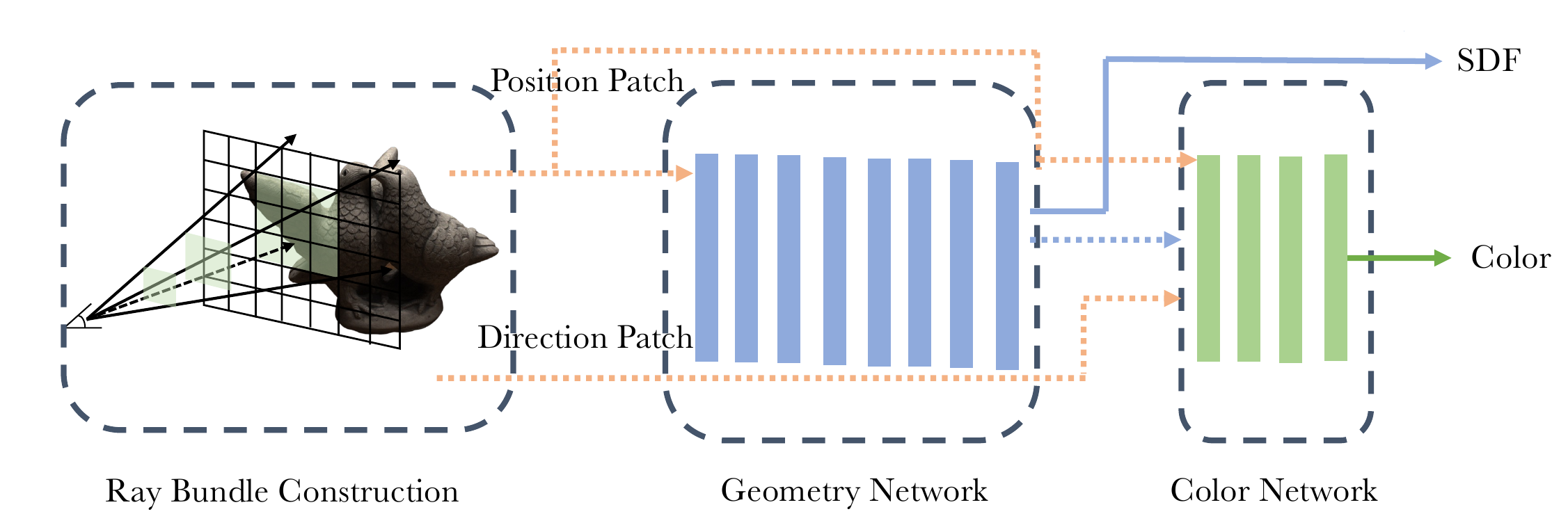}
    \caption{The BundleRecon pipeline consists of several stages. Firstly, bundle construction is performed on the input image and sampled along the rays. Secondly, the positional information is processed by the geometry network to generate the signed distance function (SDF) values and features. Finally, the direction information, along with the output of the geometry network, are fed into the color network to obtain the final color output.}
    \label{fig:model}
\end{figure*}

\subsection{Neural Implicit Multi-view Reconstruction}
Most of neural implicit multi-view reconstruction methods can be roughly categorized into two groups. The first group utilizes different implicit functions to represent the geometry while preserving the network structure of NeRF \cite{idr, unisurf, volsdf, neus}. The second category involves with designing a specialized network structure to represent complex scenes \cite{nerf-w, fig-nerf,giraffe,ocnerf}, such as those featuring foreground and background elements or static and dynamic components. In this work, we focus on the first group of method. The implicit functions utilized in these works mainly fall into two categories: the occupancy network \cite{unisurf} and the signed distance function (SDF) network \cite{idr, unisurf, volsdf, neus}. UNISURF, proposed by Oechsle \textit{et al.} \cite{unisurf}, unifies the implicit surface model and volumetric radiance model and uses the occupancy network to represent geometry. Yariv \textit{et al.} proposed IDR \cite{idr} and VolSDF \cite{volsdf}, both of which adopt the SDF network. IDR recovers the surface by disentangling geometry and appearance, while VolSDF is an improved model based on IDR. The motivation of VolSDF is to overcome the shortcomings of neural volume rendering and neural implicit surface models. Specifically, it designs a neural volume rendering model that can effectively separate geometric information and appearance. Additionally, Wang \textit{et al.} \cite{neus} present NeuS approach, which also employs the SDF network to represent geometry. NeuS represents density using the bell-shaped S-density function of SDF. Note that a bell-shaped density function is necessary to achieve an unbiased model effectively capturing thin structures \cite{neus}.

\section{Method}
\label{sec:method}

For a single pixel, the conventional approaches \cite{idr,unisurf,volsdf,neus} make use of a single ray to sample and perform neural rendering, which usually neglects the information around the pixel. In order to utilize the information from neighboring pixels, we propose a ray bundle-based method, named as BundleRecon, which is an improved sampling module that can be plugin into many existing networks. 

In this section, we firstly introduce the pipeline of ray bundle-based neural reconstruction in \ref{subsec:neurecon}. Then, we show how to construct a ray bundle in \ref{subsec:raybundle}. The ray bundle-based loss functions will be discussed in \ref{subsec:lossfunction}. Finally, the implementation details are introduced in \ref{subsec:optimization}.

\subsection{Ray Bundle Based Neural Reconstruction}
\label{subsec:neurecon}

As shown in Fig.~\ref{fig:model}, BundleRecon is composed of three parts, including the ray bundle construction, the geometry network, and the color network. Given multi-view RGB images, the first step is to construct the ray bundle. To achieve this, we randomly select patches from each image to form the ray bundle $p\in P$. Further details on the construction of the ray bundle can be found in Section \ref{subsec:raybundle}. Once the ray bundle is constructed, we can either take surface samples using methods such as \cite{idr} or conduct dense sampling around the surface as in \cite{volsdf,neus}. The positional information of the sampled points $x\in X_p$ is then fed into the geometry network, where $X_p$ is the set of sampled points that form the ray bundle. Similarly, $V_p$ is the set of directions of the rays that form the ray bundle.

The geometry network is defined by the signed distance function (SDF) $f_g: \mathbb{R}^3 \rightarrow \mathbb{R}$, which maps the spatial position of a point $x\in \mathbb{R}^3$ to its signed distance from the object. As a result, the surface $S$ of the object is represented by the zero-level-set of the geometry network, which can be expressed as follows
\begin{equation}
S=\{ x | f_g(x)=0\}.
\end{equation}

The output of the geometry network, along with the direction information $v\in V_p$, is fed into the color network $f_c: \mathbb{R}^3\times \mathbb{R}^3 \rightarrow \mathbb{R}^3$. This network maps the spatial position $x\in \mathbb{R}^3$ and the ray direction $v\in \mathbb{R}^3$, $||v||=1$, to the color patch $f_c(p)$. In other words, given a specific point in space and the direction in which the ray travels, the color network produces a corresponding color value that is used to generate the final rendered color.

To obtain the rendered color, we perform volume rendering using the following equation,
\begin{equation}
    \hat{C}(p)=\int_{0}^{\infty} w(t)f_c(p) d t,
\end{equation}
where $w(t)$ is the weight of the sampled point color. 

There are various options for selecting the weights $w(t)$. We can adopt the unbiased and occlusion-aware weights designed by NeuS\cite{neus}, or we can use regular weights in NeRF\cite{nerf}. To compute the weights, we firstly use a transformed SDF to represent the density at the sampled points, which is given by $\sigma(x)=\alpha\cdot \rho(f_g(x))$. $\alpha$ is a hyperparameter and $\rho(\cdot)$ is a transformation function, such as the cumulative distribution function of the Laplace distribution in VolSDF\cite{volsdf}. Once the density is defined, we can compute the probability of light traveling from the origin of the camera to the sampled point $x(t)$ without bouncing off, which is given by 
\begin{equation}
    T(t)=\exp \left(-\int_0^t \sigma(\boldsymbol{x}(s)) d s\right).
\end{equation}
Therefore, the probability of light being reflected at sampled point $x(t)$ and emitting color is defined as below
\begin{equation}
    O(t)=1-T(t).
\end{equation}
Thus, the weight of the sampled point color can be expressed as follows
\begin{equation}
    w(t)=\frac{d O}{d t}(t)=\sigma(\boldsymbol{x}(t)) T(t).
\end{equation}

In addition to the volume rendering mentioned above, BundleRecon can also use rendering methods in IDR \cite{idr}.
\begin{equation}
\small
    \hat{C}(p)=L^e\left(\hat{x}, v^o\right)+\int_{\Omega} B\left(\hat{x}, \hat{n}, v^i, v^o\right) L^i\left(\hat{x},v^i\right)\left(\hat{n} \cdot v^i\right) d v^i ,
\end{equation}
where $B$ refers to the bidirectional reflectance distribution function (BRDF). $\hat{x}$ is the surface point obtained by ray tracing. $\hat{n}$ is the normal vector computed by SDF and $\Omega$ is half sphere centered at $\hat{n}$. $v^i$ and $v^o$ represent the incoming and outgoing direction of the ray, respectively. 

In summary, our model is compatible with the sampling and rendering methods for most of existing implicit models, such as NeRF \cite{nerf}, IDR \cite{idr}, VolSDF \cite{volsdf}, NeuS \cite{neus} and UNISURF \cite{unisurf}.

\subsection{Ray Bundle}
\label{subsec:raybundle}

We randomly select $n$ pixels from the input image like the traditional methods \cite{nerf,idr,volsdf,neus,unisurf}. Instead of simply considering each pixel in isolation, we employ it as the anchor of an $s \times s$ pixel patch, which is named as a ray bundle. Each ray bundle consists of $s^2$ pixels, and we construct $n$ such bundles for a total of $ns^2$ pixels. This is in contrast to traditional methods that only consider $n$ rays with $n$ pixels. As a result, our approach incurs a higher memory cost. To mitigate this, we may reduce the number of ray bundles used, however, this would inevitably compromise the quality of rendering and reconstruction, necessitating a trade-off. To reduce memory usage and enhance efficiency, when integrating BundleRecon into models \cite{volsdf,neus,nerf} that have numerous sample points on a single ray, we can employ dense sampling for the central ray of each ray bundle. Moreover, the sparse sampling is used for the surrounding rays. In contrast, models \cite{idr} that rely on ray tracing to obtain the surface point as the only sample point on each ray can readily incorporate BundleRecon without any changes to their sampling strategy. 

\begin{figure}[htbp]
    \centering
    \includegraphics[width=8cm]{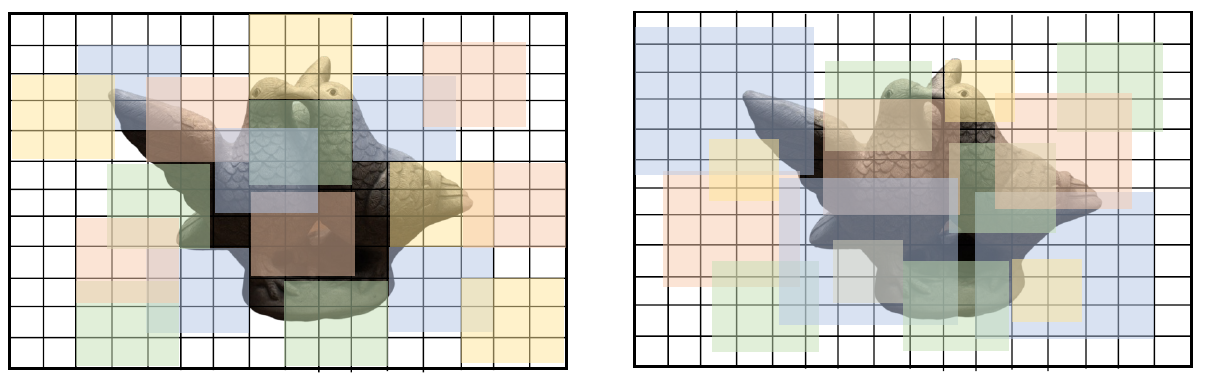}
    \caption{Each square in a different color represents a distinct epoch in the training process. The left figure indicates a fixed bundle size throughout the training, whereas the right figure depicts a varying bundle size during the training process.}
    \label{fig:construction}
\end{figure}

As mentioned above, the ray bundle itself is a regular square patch. Fig.~\ref{fig:construction} illustrates two ways of constructing a square bundle: a fixed-size bundle, and a bundle whose size changes dynamically during training. Additionally, we can construct ray bundles using super-pixels that incorporate the semantic information.

Consider using a flashlight to illuminate an object. The resulting ray bundle may cover a flat area or intersect an edge, with some parts belonging to the front of the object and others to the back. In the latter case, the information on the two sides of the edge can be quite different, and it may not be necessary to treat them as a whole. To address this issue, we partition such areas based on the spatial location of the sampling points along the ray bundle. Specifically, we compute the distance between each pixel and its eight surroundings to generate a distance mask $M_P$. During training, this mask is used to filter out the irrelevant information.

\subsection{Loss Function}
\label{subsec:lossfunction}
Similar to the traditional methods~\cite{nerf,unisurf,idr,volsdf,neus}, we employ color loss $\mathcal{L}_{c} $ to supervise the color network as well as the geometry network. For a ray bundle, we have the following loss function
\begin{equation}
    \mathcal{L}_{c} = \sum_{p\in P}||\hat{C}(p)-C(p)||_1 ,
\end{equation}
where $P$ denotes the set of all ray bundle, $\hat{C}(p)$ is the rendered color with respect to the ray bundle $p$, and $C(p)$ is the true color of the pixel patch.

To further leverage the information from neighboring pixels, we compute the mean and variance of the color patch. We denote the mean and variance operations as $M(\cdot)$ and $V(\cdot)$, respectively. The loss $\mathcal{L}_{m}$ and $\mathcal{L}_{v}$ can ben obtained as below
\begin{equation}
    \mathcal{L}_{m} = ||M(\hat{C}(P))-M(C(P))||_2 ,
\end{equation}

\begin{equation}
    \mathcal{L}_{v} = ||V(\hat{C}(P))-V(C(P))||_2 .
\end{equation}

For color patches, we can also extract their convolutional features. To process the patch information, we need to apply the distance mask $M_P$ to filter out irrelevant details from the color patches. We obtain the features of color patches by convolving them with a Sobel kernel, denoted as $F(\cdot)$. The loss function for this process is formulated as follows
\begin{equation}
    \mathcal{L}_{conv} = ||(F(\hat{C}(P))-F(C(P)))\cdot M_{P}||_2 .
\end{equation}

\subsection{Implementation Details}
\label{subsec:optimization}

Our model can be easily integrated into the existing single ray based neural implicit models, so the architecture details can remain the same. In this paper, we demonstrate the effectiveness of our model using IDR \cite{idr} and NeuS \cite{neus}. It is noteworthy that, our model can also be plugged into other models like VolSDF \cite{volsdf}. During the experiment, we set the pixel patch size to be $3 \times 3$, which remained constant during training. $n=229$ ray bundles are used for sampling, which results in a batch size of 2061. We trained our model for 2000 epochs on IDR and 100K iterations on NeuS.

The loss function used in the experiment is as follows:
\begin{equation}
\mathcal{L}=\lambda_{c}\mathcal{L}_{c}+\lambda_{m}\mathcal{L}_{m}+\lambda_{v}\mathcal{L}_{v}+\lambda_{conv}\mathcal{L}_{conv},
\end{equation}
where $\lambda_{c}=1$, $\lambda_{m}=5e^{-3}$, $\lambda_{v}=1e^{-2}$ and $\lambda_{conv}=5e^{-5}$. 

\section{Experiments}
\label{sec:experiments}

\subsection{Experimental Settings}
\textbf{Datasets:} We evaluate our model on 15 scenes from the DTU dataset\cite{dtu} and 4 scenes from the BlendedMVS dataset\cite{blendedmvs}. The DTU dataset includes multi-view images of various objects with the fixed camera and lighting parameters. Each image has a resolution of $1600\times 1200$, and DTU also provides masks and real point cloud data, which facilitates mask processing during training and evaluation of the Chamfer distance of the reconstructed mesh. For the BlendedMVS dataset, masks and camera parameters are also provided. Their scenes have more complex backgrounds compared to DTU.

\textbf{Baseline.} We first integrate our proposed module into IDR~\cite{idr} and use it as a baseline to evaluate the effectiveness of BundleRecon. Due to memory limitation, for each image, we used 229 ray bundles for the experiment. To ensure the fairness of the experiment, we lowered the number of sampled pixels of IDR to 229 for comparison as well. Besides, to validate the compatibility of BundleRecon, we integrate it into NeuS \cite{neus} to perform quantitative comparision. The number of the sampled pixels in NeuS is set to 299, and the number of ray bundle is set to 114. In this section, we use IDR* and NeuS* to represent our different settings from the original model.

\subsection{Multi-view 3D Reconstruction}
\label{subsec:reconstruction}

\begin{table*}[htb]
\caption{Quantitative Comparison Results}
\label{table:vs}
\centering
\begin{tabular}{c|cc|cc||cc|cc}
\hline
\multicolumn{1}{l|}{\textbf{}} & \multicolumn{2}{c|}{\textbf{IDR*}} & \multicolumn{2}{c||}{\textbf{IDR+BundleRecon}} & \multicolumn{2}{c|}{\textbf{NeuS*}} & \multicolumn{2}{c}{\textbf{NeuS+BundleRecon}} \\ \cline{2-9} 
\textbf{scan}                  & \textbf{Chamfer}  & \textbf{PSNR}  & \textbf{Chamfer}   & \textbf{PSNR}   & \textbf{Chamfer}  & \textbf{PSNR}   & \textbf{Chamfer}   & \textbf{PSNR}   \\ \hline
24                             & 2.03              & 21.99          & \textbf{1.82}      & \textbf{22.36}  & 1.42              & 22.98           & \textbf{1.31}      & \textbf{24.41}  \\
37                             & \textbf{2.02}     & 19.50          & 2.12               & \textbf{19.95}  & 1.58              & 19.15           & \textbf{1.45}      & \textbf{21.99}  \\
40                             & 0.87              & 23.81          & \textbf{0.85}      & \textbf{24.32}  & 1.59              & 23.16           & \textbf{1.18}      & \textbf{25.41}  \\
55                             & 0.52              & 20.33          & \textbf{0.45}      & \textbf{21.63}  & 0.62              & \textbf{20.94}  & \textbf{0.54}      & 20.65           \\
63                             & 1.59              & 21.79          & \textbf{1.31}      & \textbf{22.58}  & \textbf{1.79}     & 25.17           & 1.81               & \textbf{25.36}  \\
65                             & 1.05              & \textbf{23.42} & \textbf{1.03}      & 23.33           & \textbf{0.87}     & \textbf{28.15}  & 0.88               & 27.90           \\
69                             & 0.93              & \textbf{20.84} & \textbf{0.90}      & 20.65           & \textbf{0.70}     & \textbf{25.44}  & 0.79               & 24.84           \\
83                             & \textbf{1.29}     & 19.63          & 1.48               & \textbf{24.32}  & 1.59              & 27.02           & \textbf{1.16}      & \textbf{30.91}  \\
97                             & \textbf{1.46}     & 21.91          & 1.55               & \textbf{22.20}  & 1.46              & \textbf{23.87}  & \textbf{1.45}      & 23.65           \\
105                            & 0.73              & 20.96          & \textbf{0.70}      & \textbf{22.41}  & 1.29              & 25.29           & \textbf{1.09}      & \textbf{27.30}  \\
106                            & \textbf{0.74}     & \textbf{20.77} & 0.77               & 20.28           & 0.69              & 31.04           & \textbf{0.64}      & \textbf{31.90}  \\
110                            & 1.41              & 20.64          & \textbf{1.29}      & \textbf{20.88}  & 2.05              & 26.54           & \textbf{1.81}      & \textbf{27.02}  \\
114                            & 0.56              & 23.44          & \textbf{0.38}      & \textbf{25.34}  & 0.47              & 26.12           & \textbf{0.42}      & \textbf{26.34}  \\
118                            & 0.65              & 21.75          & \textbf{0.58}      & \textbf{22.96}  & 0.62              & 31.27           & \textbf{0.57}      & \textbf{32.66}  \\
122                            & 0.60              & 25.81          & \textbf{0.57}      & \textbf{25.97}  & \textbf{0.63}     & \textbf{30.95}  & 0.68               & 27.60           \\ \hline
mean                           & 1.10              & 21.77          & \textbf{1.05}      & \textbf{22.61}  & 1.16              & 25.81           & \textbf{1.05}      & \textbf{26.53}  \\ \hline
\end{tabular}
\end{table*}

\begin{figure}[htbp]
    \centering
    \includegraphics[width=5.3cm]{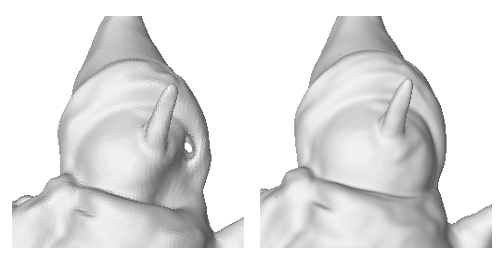}
    \caption{Comparison of reconstruction details. There are holes in the mesh generated by IDR* (left), and the correct geometry can be obtained by integrating BundleRecon (right).}
    \label{fig:vs_hole}
\end{figure}

\begin{figure*}[htbp]
\centering     

\begin{minipage}{0.04\linewidth}
    \vspace{3pt}
    \rotatebox{90}{Reference Images}
\end{minipage}
\begin{minipage}{0.23\linewidth}
    \vspace{3pt}
    \centerline{\includegraphics[width=\textwidth]{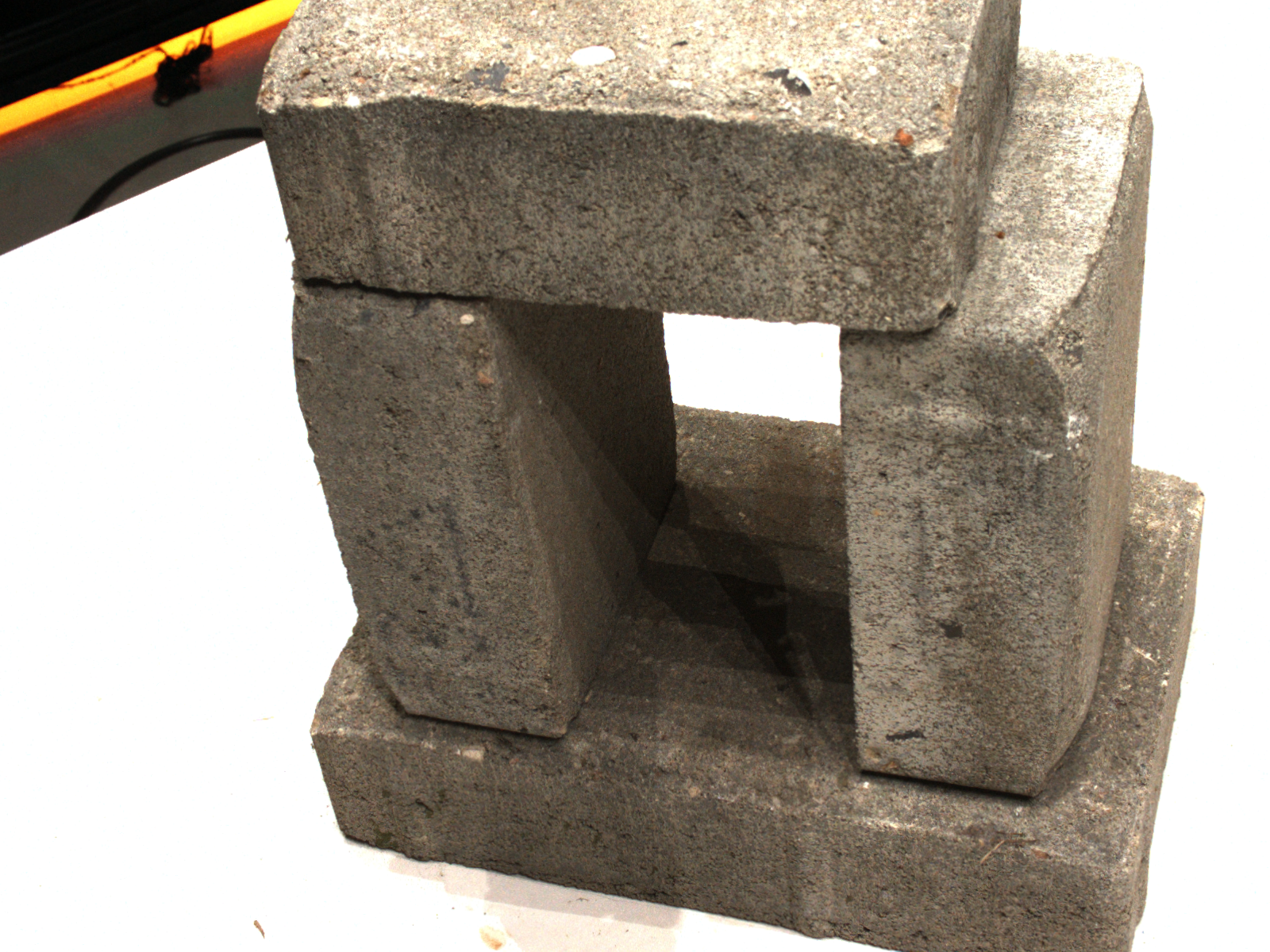}}
\end{minipage}
\begin{minipage}{0.23\linewidth}
    \vspace{3pt}
    \centerline{\includegraphics[width=\textwidth]{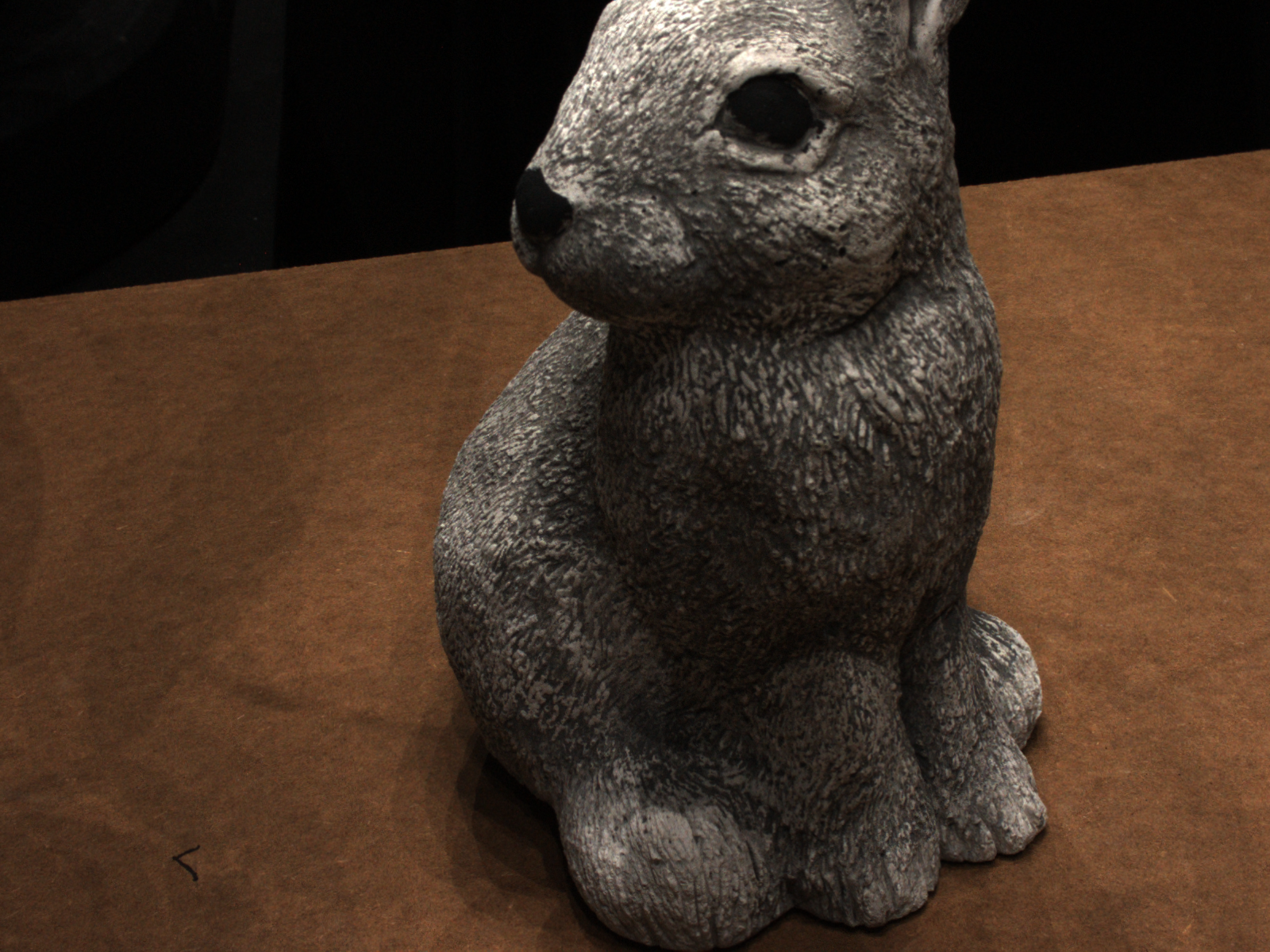}}
\end{minipage}
\begin{minipage}{0.23\linewidth}
    \vspace{3pt}
    \centerline{\includegraphics[width=\textwidth]{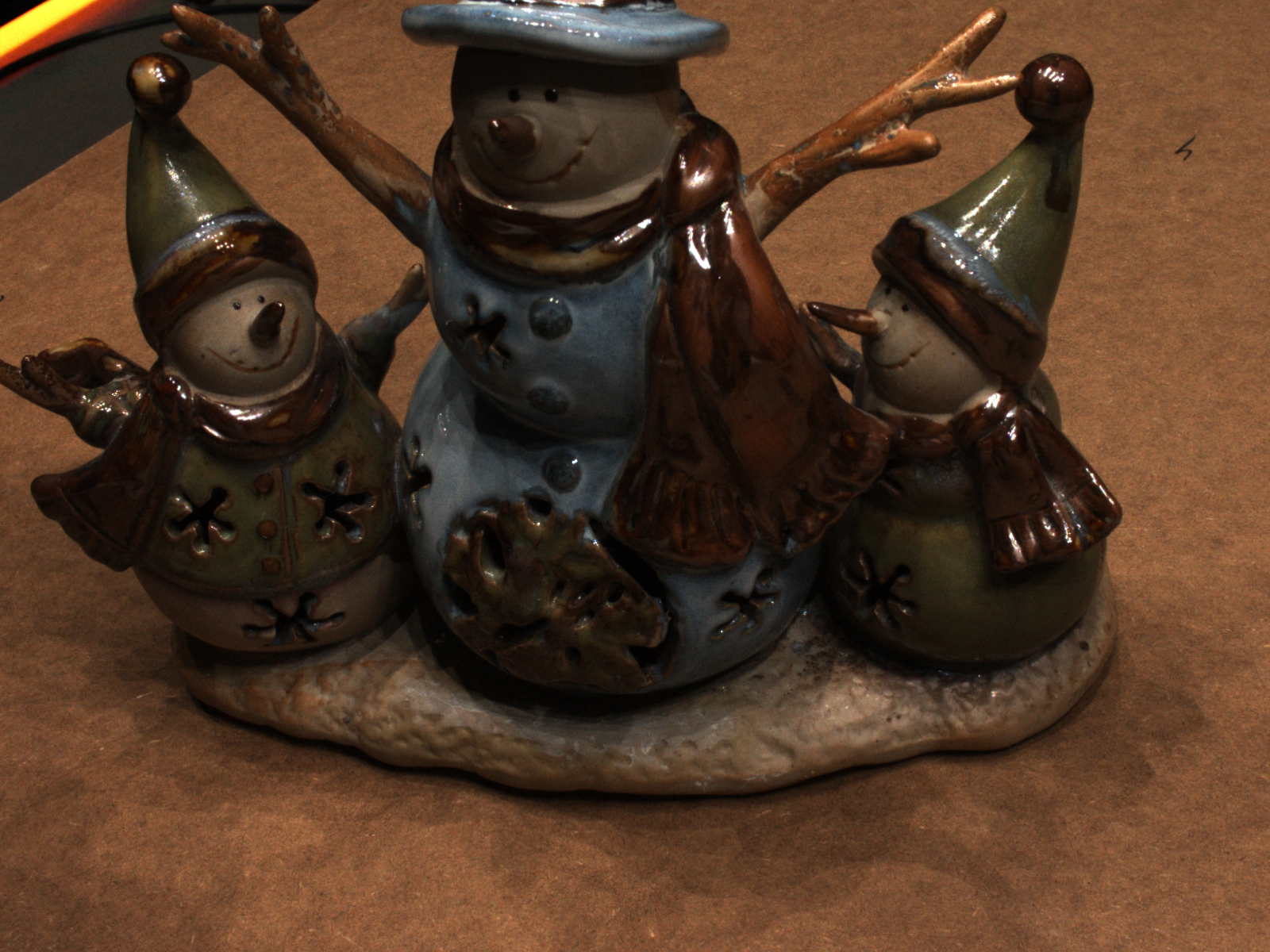}}
\end{minipage}
\begin{minipage}{0.23\linewidth}
    \vspace{3pt}
    \centerline{\includegraphics[width=\textwidth]{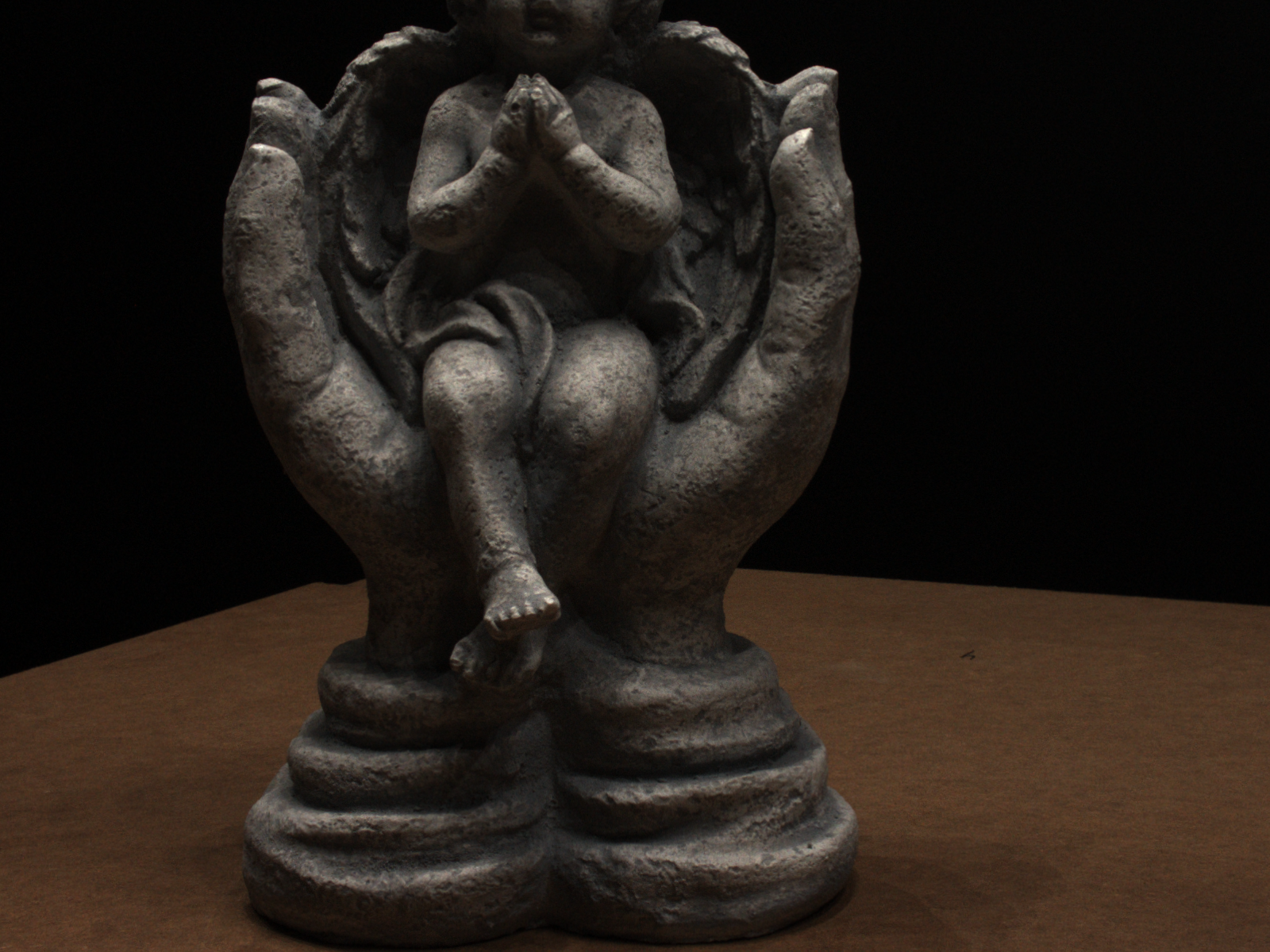}}
\end{minipage}
  
\begin{minipage}{0.04\linewidth}
    \vspace{3pt}
    \rotatebox{90}{IDR*}
\end{minipage}
\begin{minipage}{0.23\linewidth}
    \vspace{3pt}
    \centerline{\includegraphics[width=\textwidth]{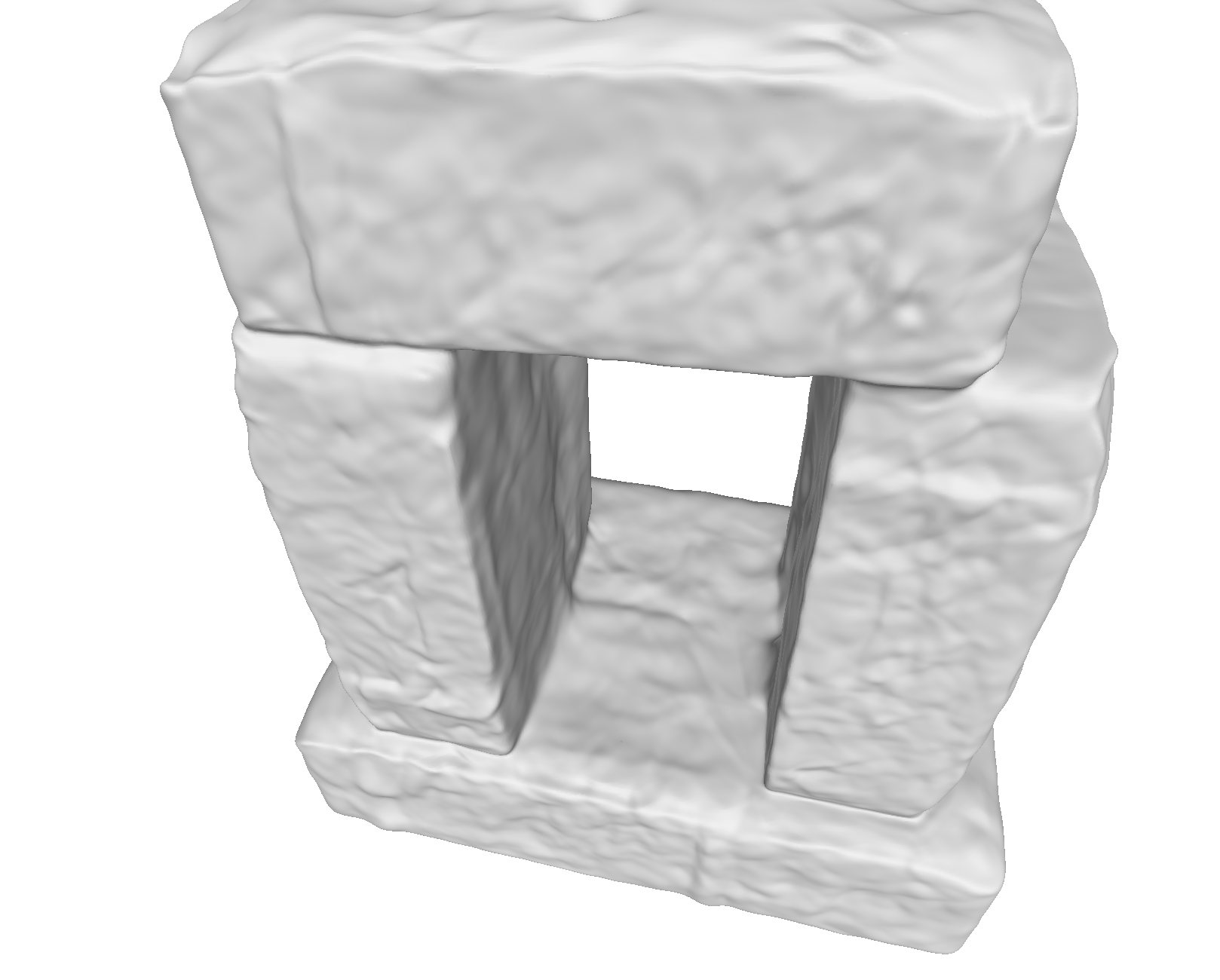}}
\end{minipage}
\begin{minipage}{0.23\linewidth}
    \vspace{3pt}
    \centerline{\includegraphics[width=\textwidth]{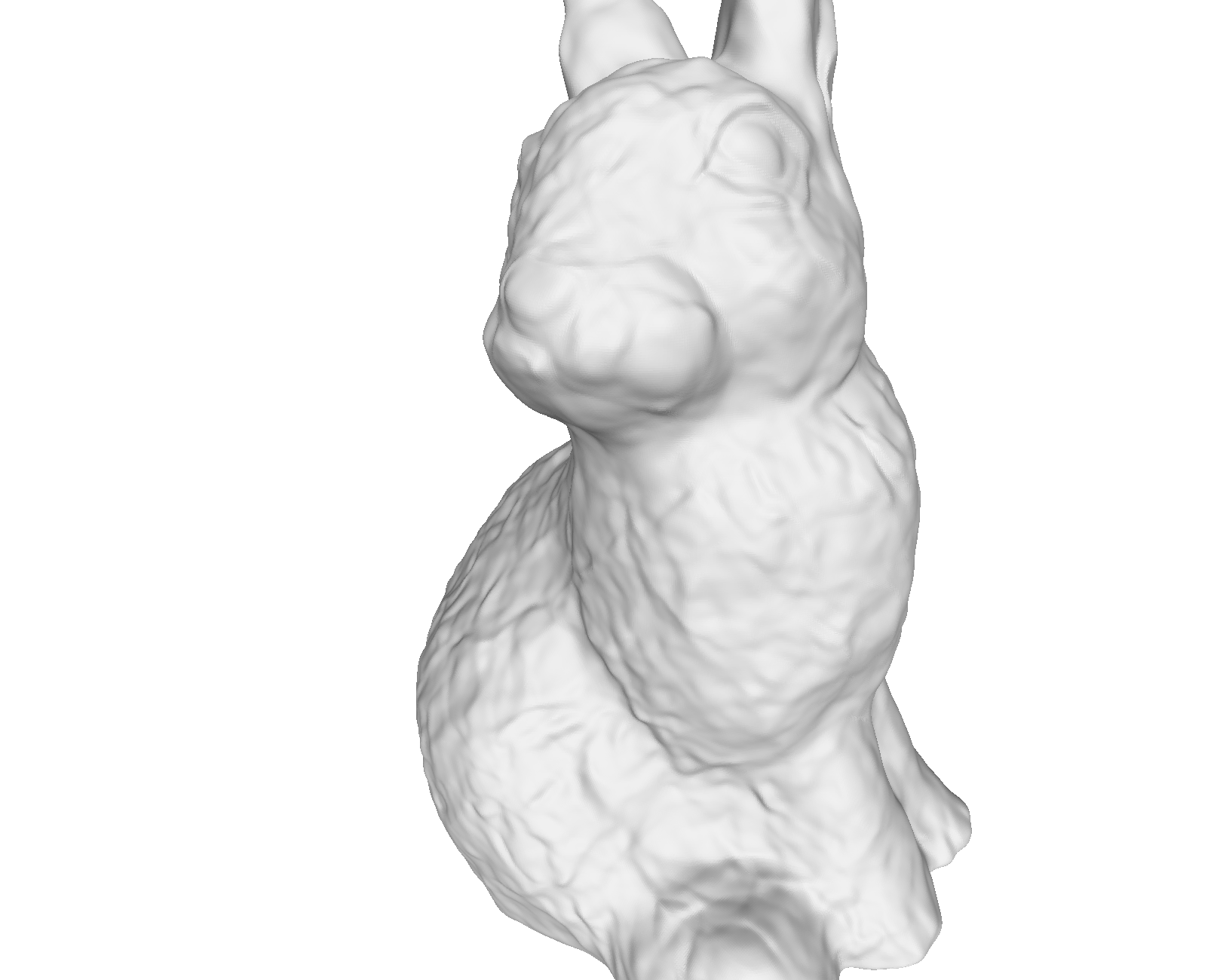}}
\end{minipage}
\begin{minipage}{0.23\linewidth}
    \vspace{3pt}
    \centerline{\includegraphics[width=\textwidth]{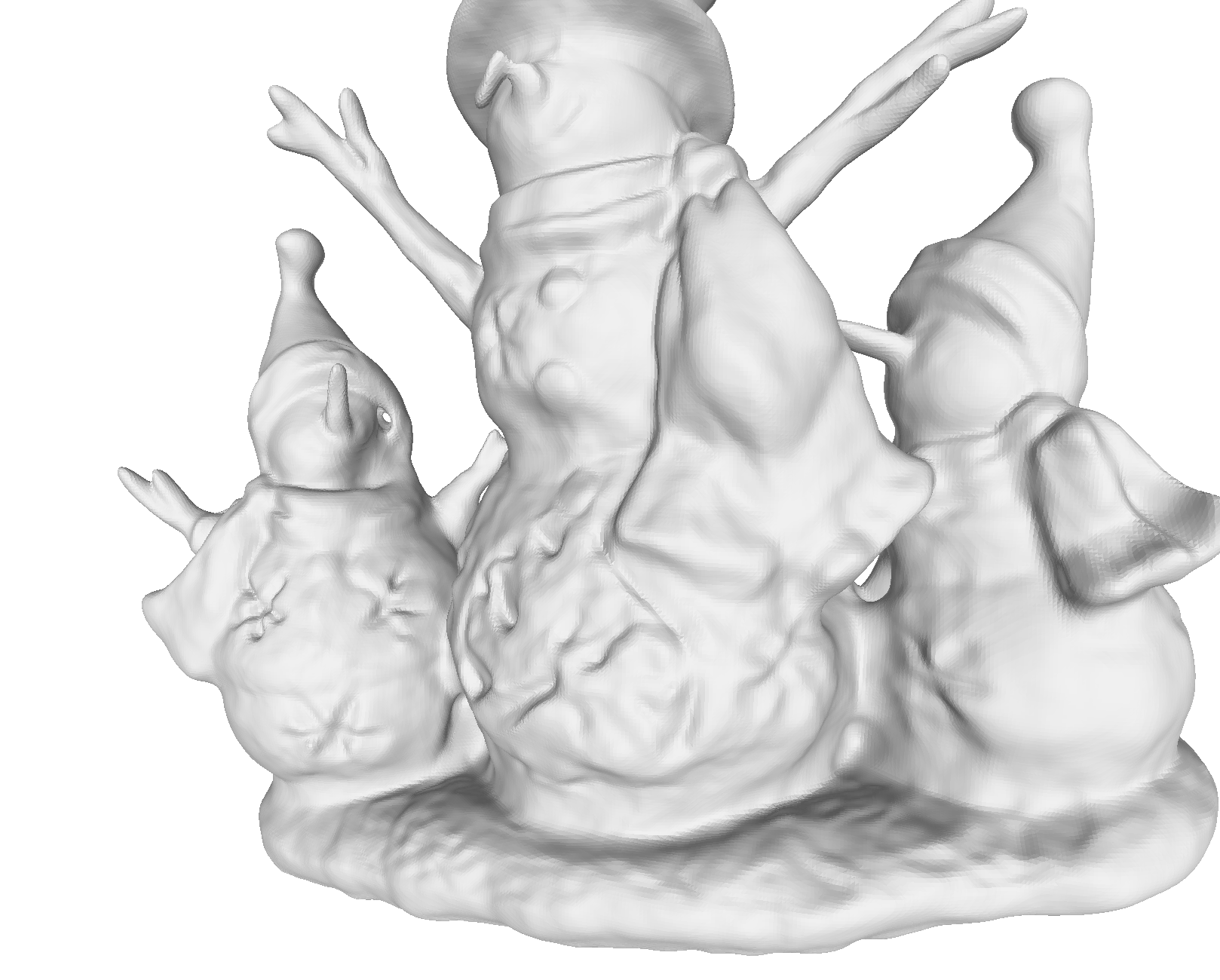}}
\end{minipage}
\begin{minipage}{0.23\linewidth}
    \vspace{3pt}
    \centerline{\includegraphics[width=\textwidth]{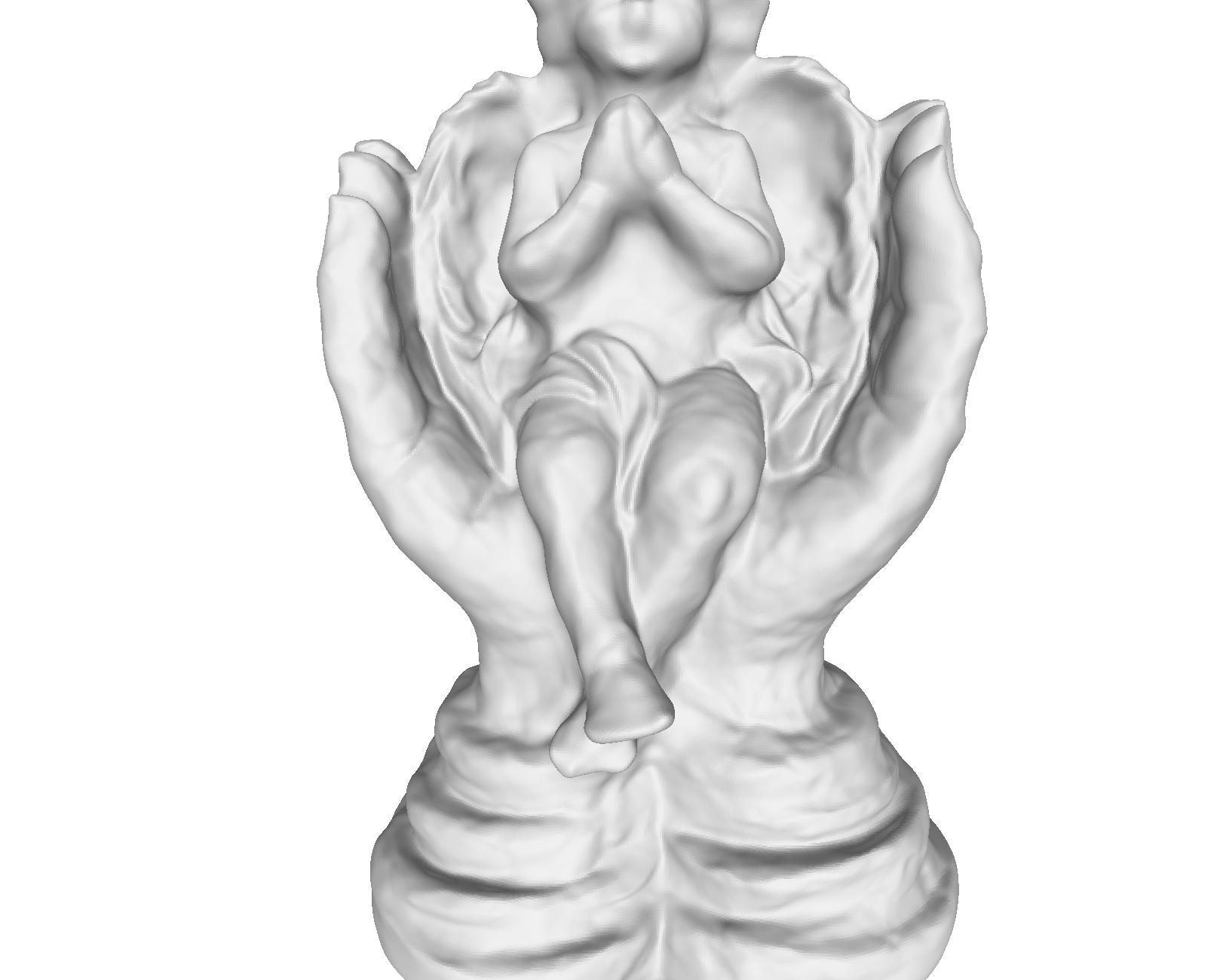}}
\end{minipage}

\begin{minipage}{0.04\linewidth}
    \vspace{3pt}
    \rotatebox{90}{IDR+BundleRecon}
\end{minipage}
\begin{minipage}{0.23\linewidth}
    \vspace{3pt}
    \centerline{\includegraphics[width=\textwidth]{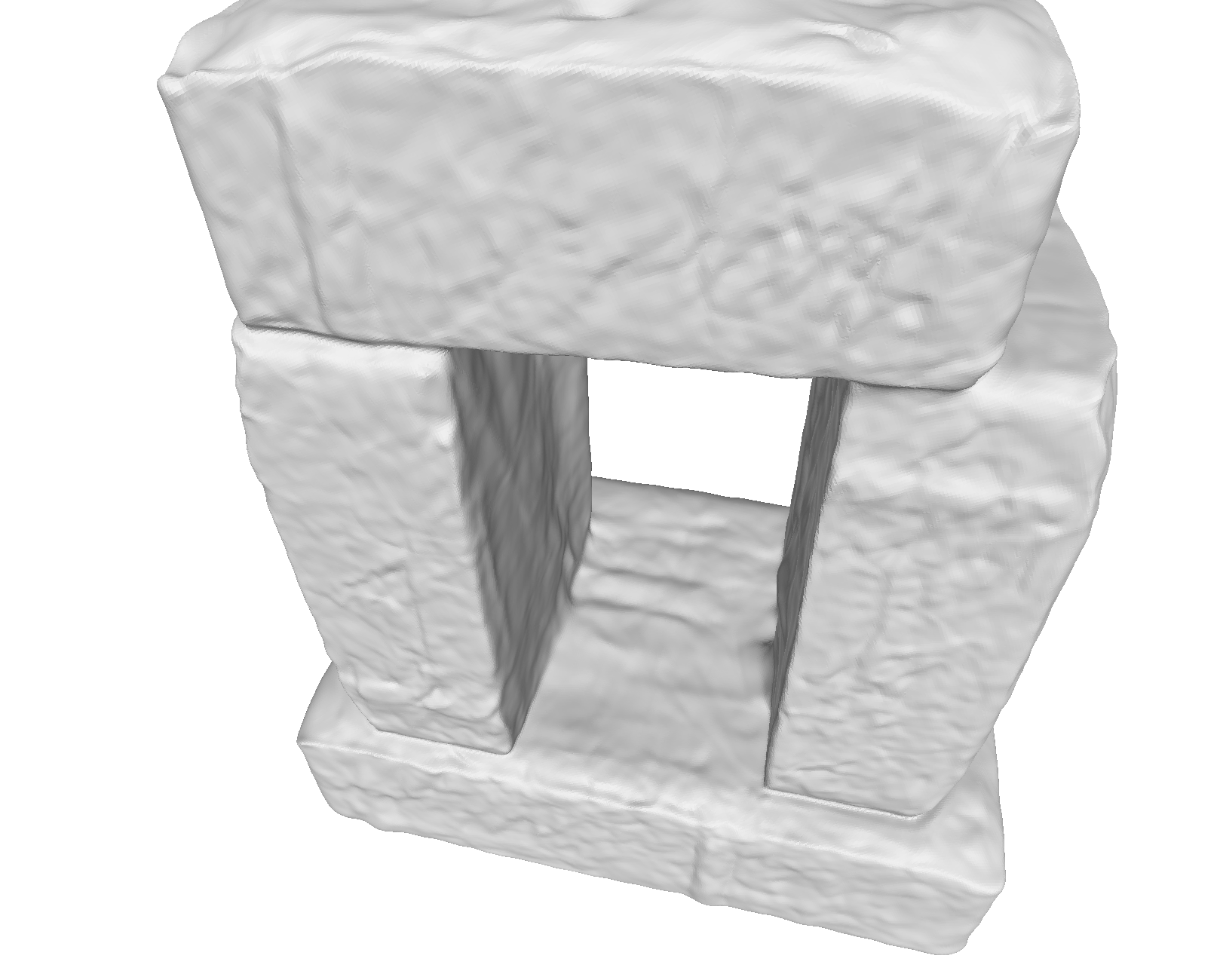}}
\end{minipage}
\begin{minipage}{0.23\linewidth}
    \vspace{3pt}
    \centerline{\includegraphics[width=\textwidth]{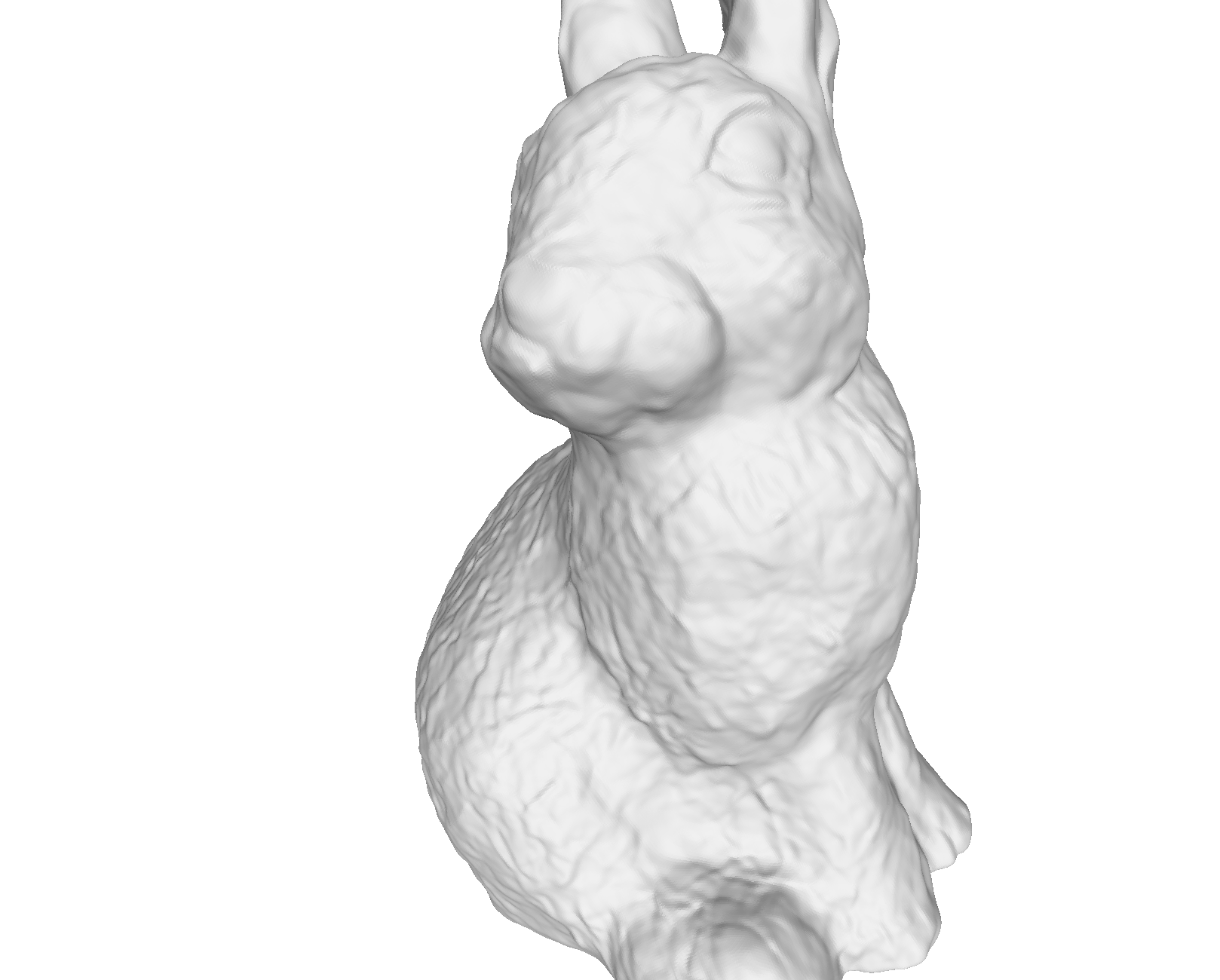}}
\end{minipage}
\begin{minipage}{0.23\linewidth}
    \vspace{3pt}
    \centerline{\includegraphics[width=\textwidth]{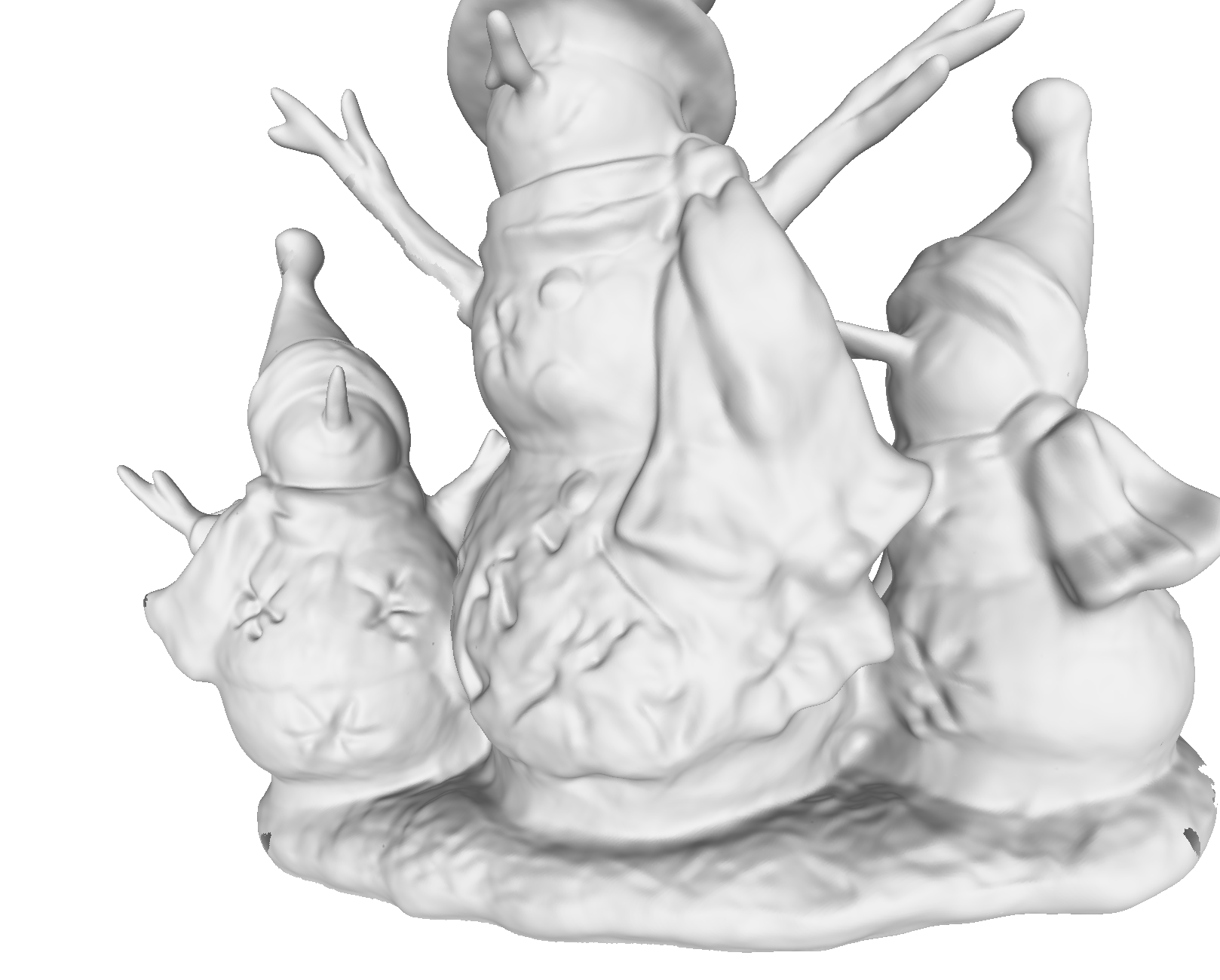}}
\end{minipage}
\begin{minipage}{0.23\linewidth}
    \vspace{3pt}
    \centerline{\includegraphics[width=\textwidth]{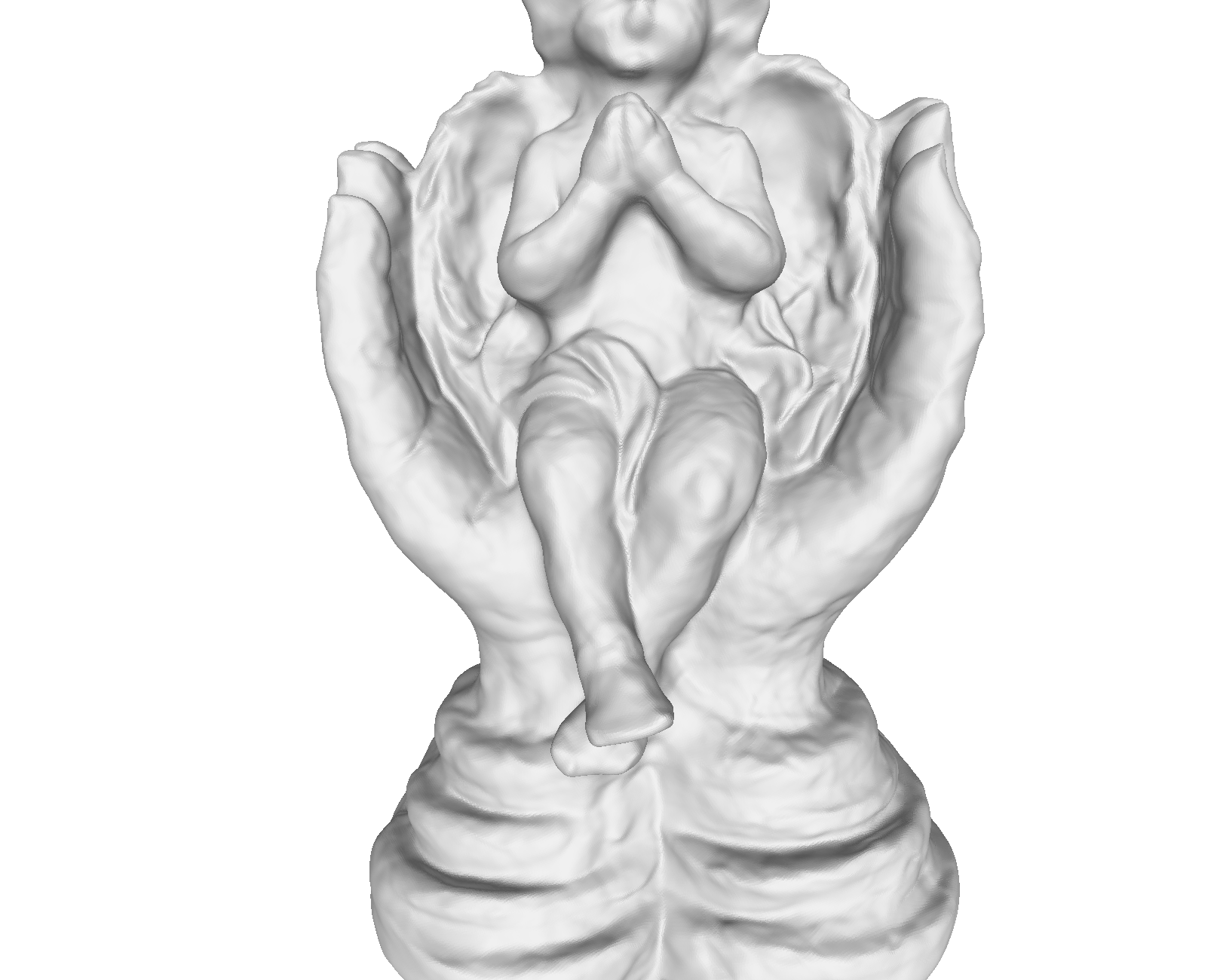}}
\end{minipage}

\caption{Qualitative comparison of IDR* and BundleRecon on DTU}
\label{fig:qualitative}
\end{figure*}

The qualitative results of IDR+BundleRecon is presented in Fig.~\ref{fig:qualitative}. It is evident that the mesh generated by IDR+BundleRecon captures more details. The first and the second scenes demonstrate that the mesh generated by our model has more refined geometric textures. In the third scene, the mesh generated by IDR+BundleRecon eliminates the geometric errors in the snowman's face. In the fourth scene, the mesh generated by IDR+BundleRecon exhibits better performance in the area around the arm. One of the reconstruction details is shown in Fig.~\ref{fig:vs_hole}. Traditional models such as IDR require sampling as many pixels as possible to obtain the object details, which may lead to the incorrect geometry when the number of sampled pixels is reduced, as seen in the hole on the left. Fortunately, the incorporation of BundleRecon can effectively alleviate this issue.

\begin{figure*}[htbp]
\centering     

\begin{minipage}{0.04\linewidth}
    \vspace{3pt}
    \rotatebox{90}{Reference Images}
\end{minipage}
\begin{minipage}{0.23\linewidth}
    \vspace{3pt}
    \centerline{\includegraphics[width=\textwidth]{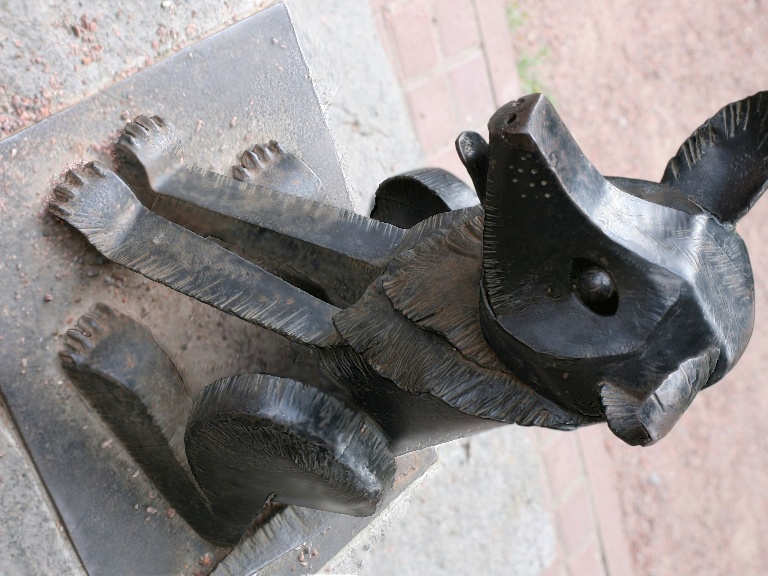}}
\end{minipage}
\begin{minipage}{0.23\linewidth}
    \vspace{3pt}
    \centerline{\includegraphics[width=\textwidth]{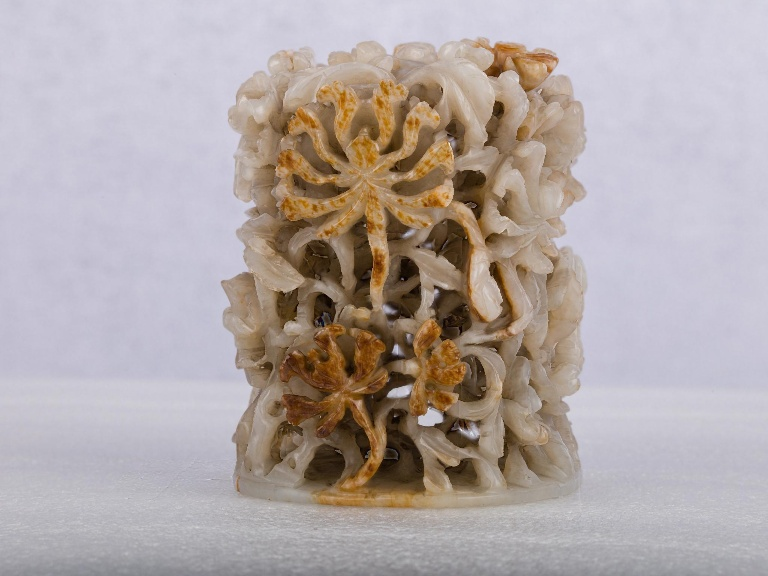}}
\end{minipage}
\begin{minipage}{0.23\linewidth}
    \vspace{3pt}
    \centerline{\includegraphics[width=\textwidth]{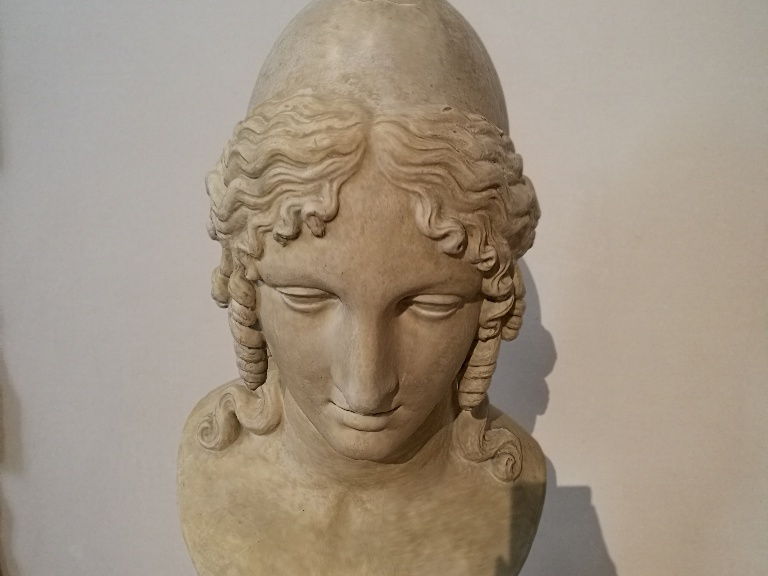}}
\end{minipage}
\begin{minipage}{0.23\linewidth}
    \vspace{3pt}
    \centerline{\includegraphics[width=\textwidth]{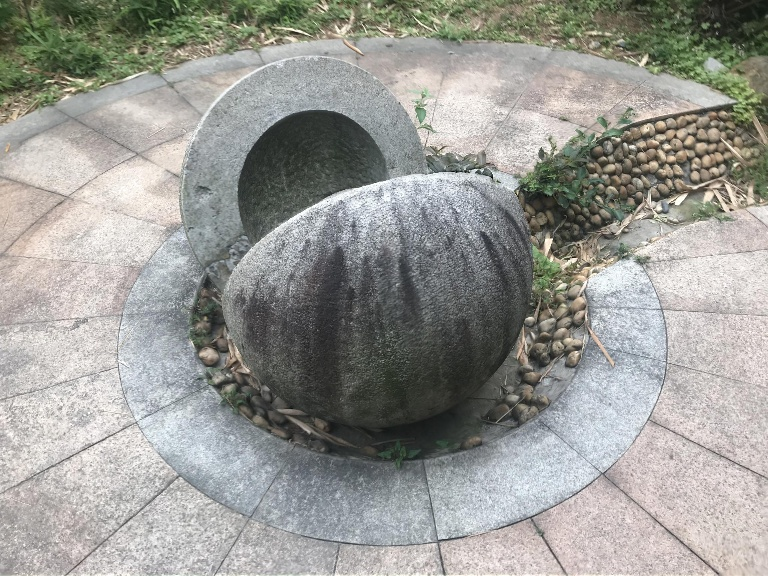}}
\end{minipage}
  
\begin{minipage}{0.04\linewidth}
    \vspace{3pt}
    \rotatebox{90}{NeuS*}
\end{minipage}
\begin{minipage}{0.23\linewidth}
    \vspace{3pt}
    \centerline{\includegraphics[width=\textwidth]{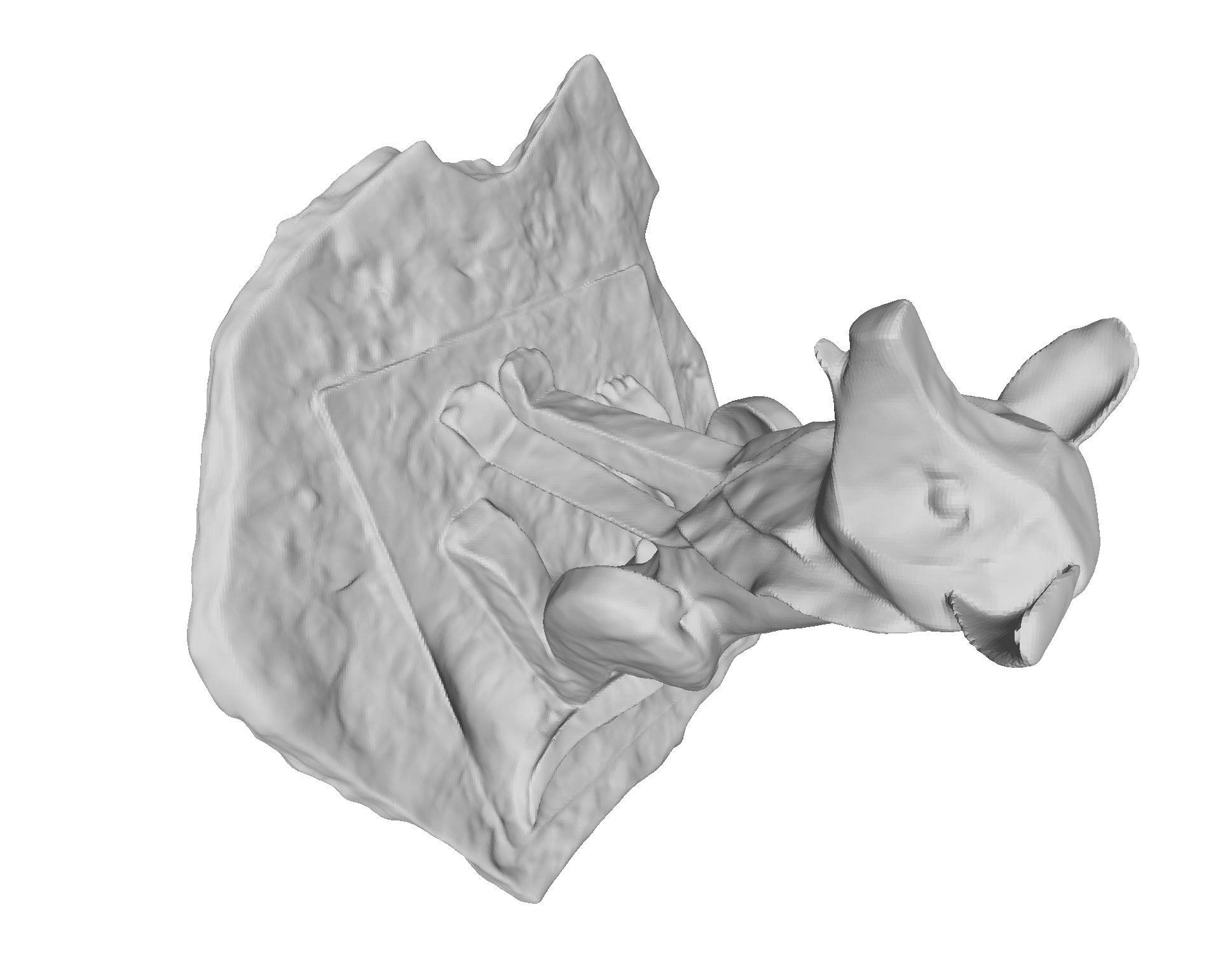}}
\end{minipage}
\begin{minipage}{0.23\linewidth}
    \vspace{3pt}
    \centerline{\includegraphics[width=\textwidth]{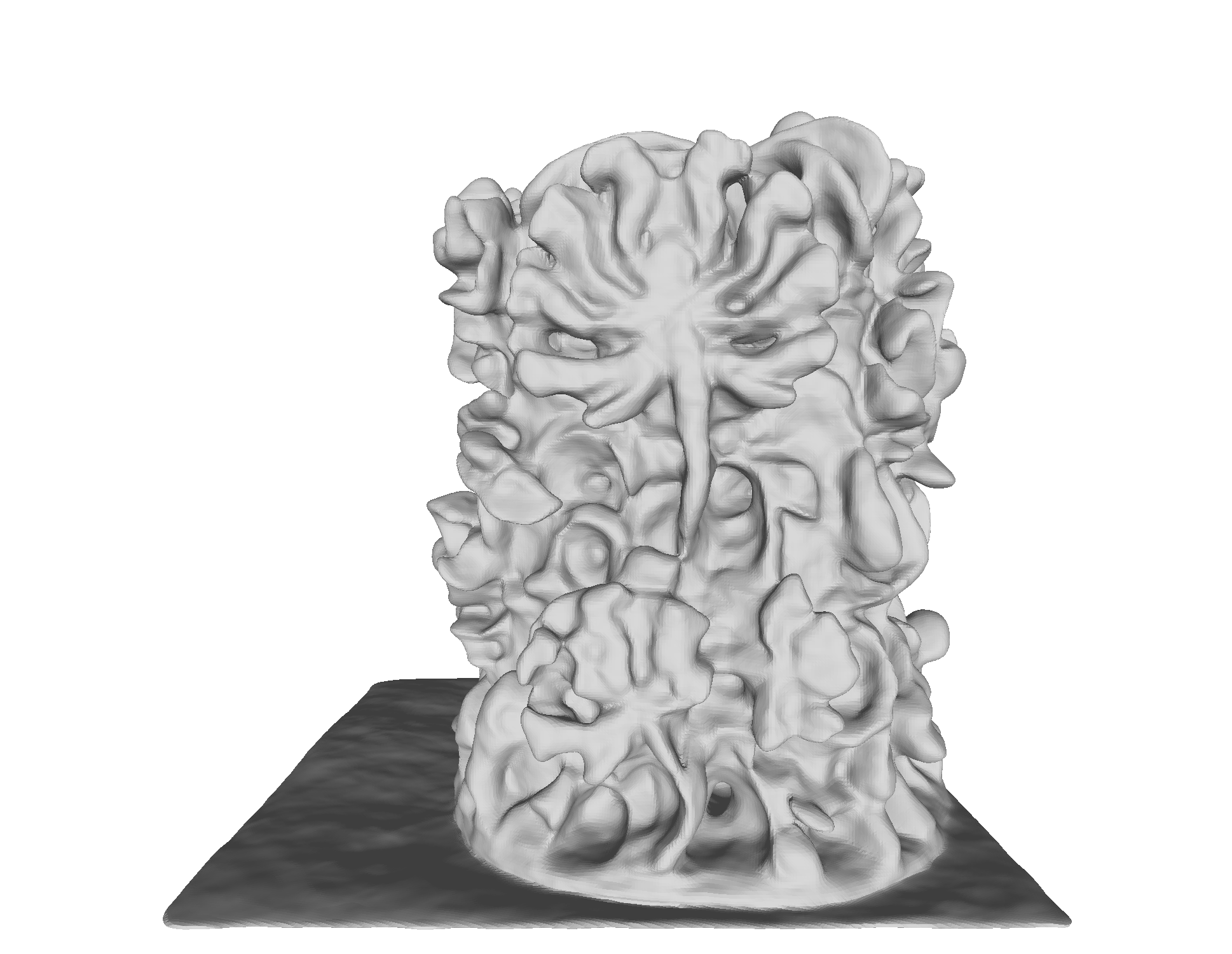}}
\end{minipage}
\begin{minipage}{0.23\linewidth}
    \vspace{3pt}
    \centerline{\includegraphics[width=\textwidth]{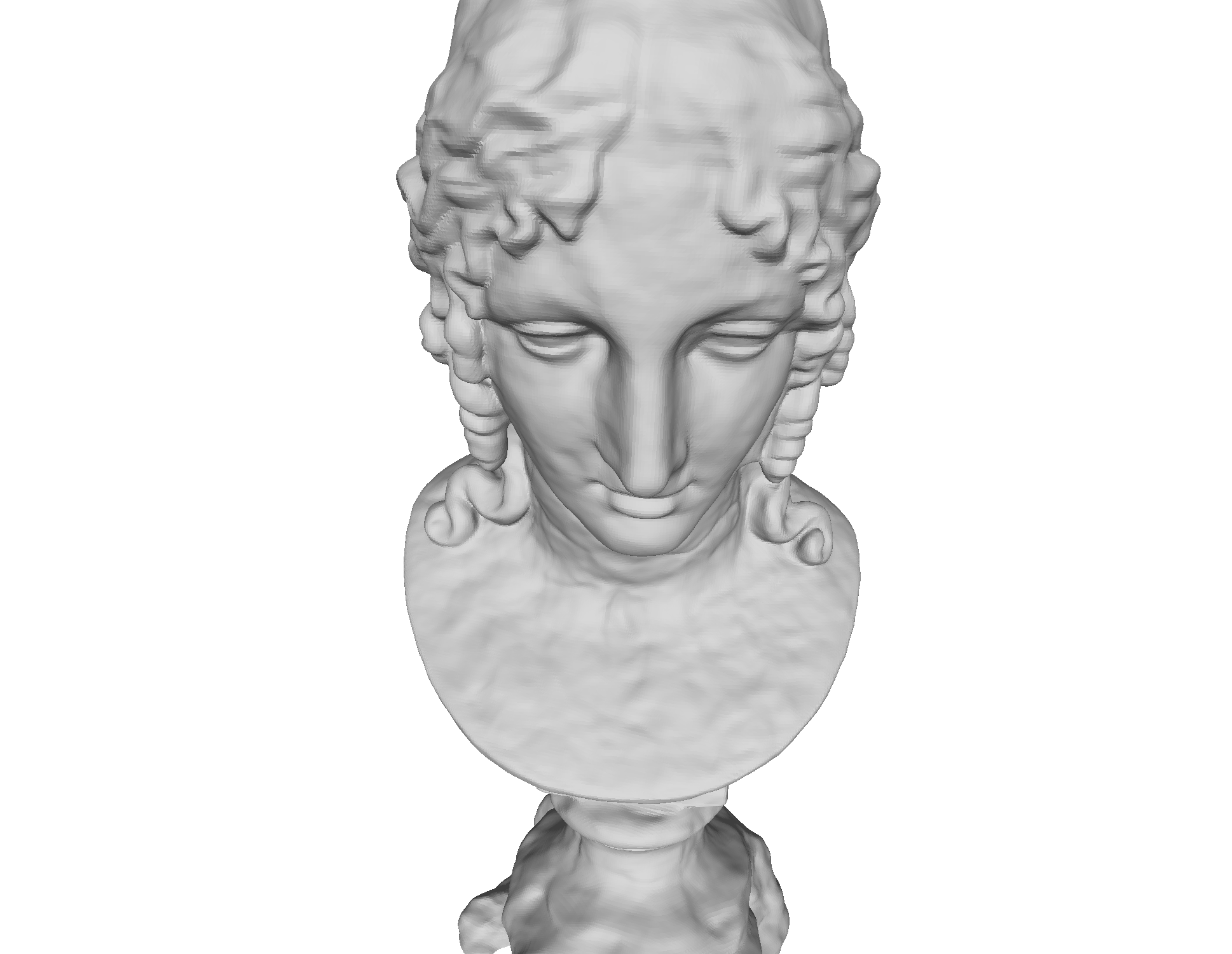}}
\end{minipage}
\begin{minipage}{0.23\linewidth}
    \vspace{3pt}
    \centerline{\includegraphics[width=\textwidth]{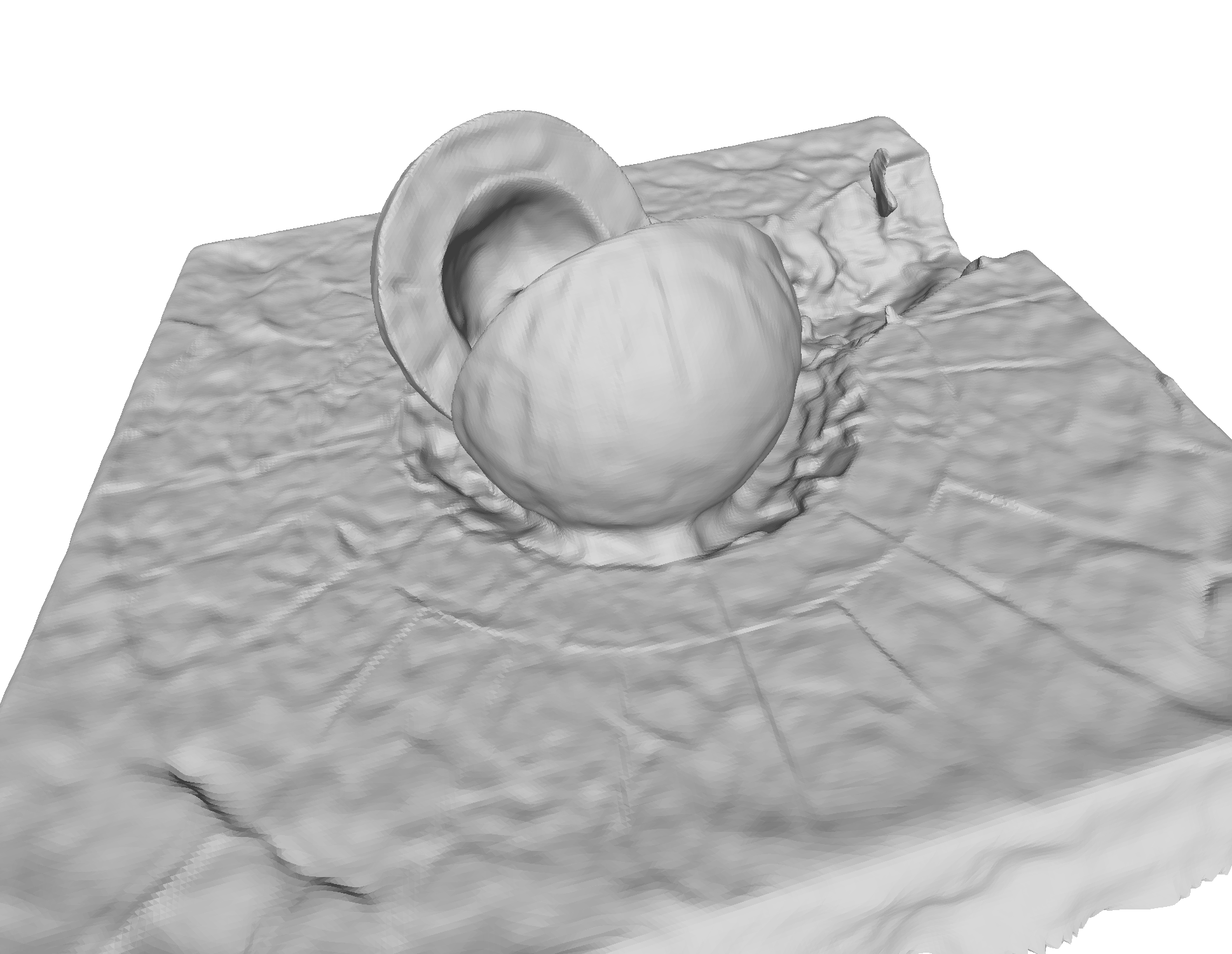}}
\end{minipage}

\begin{minipage}{0.04\linewidth}
    \vspace{3pt}
    \rotatebox{90}{NeuS+BundleRecon}
\end{minipage}
\begin{minipage}{0.23\linewidth}
    \vspace{3pt}
    \centerline{\includegraphics[width=\textwidth]{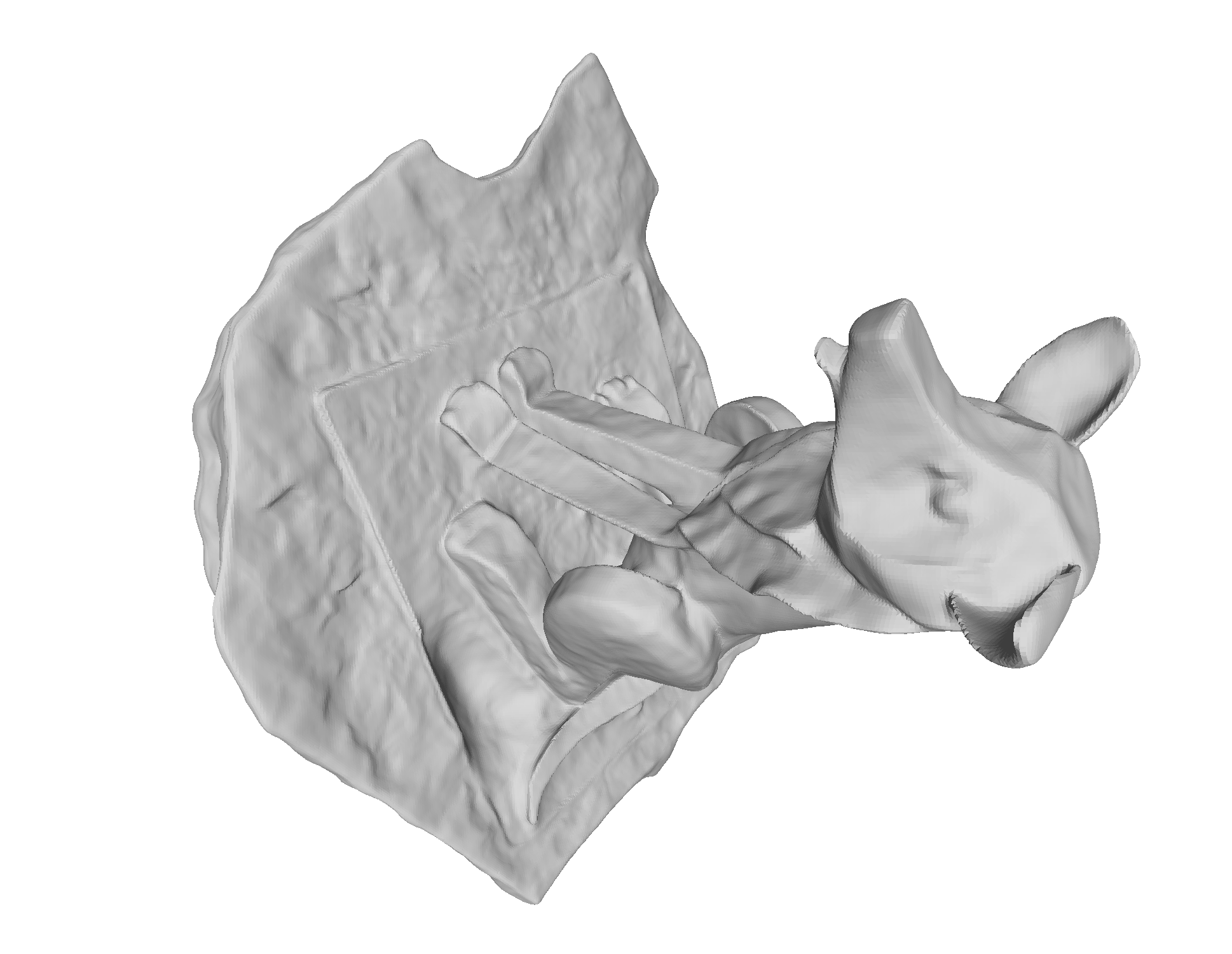}}
\end{minipage}
\begin{minipage}{0.23\linewidth}
    \vspace{3pt}
    \centerline{\includegraphics[width=\textwidth]{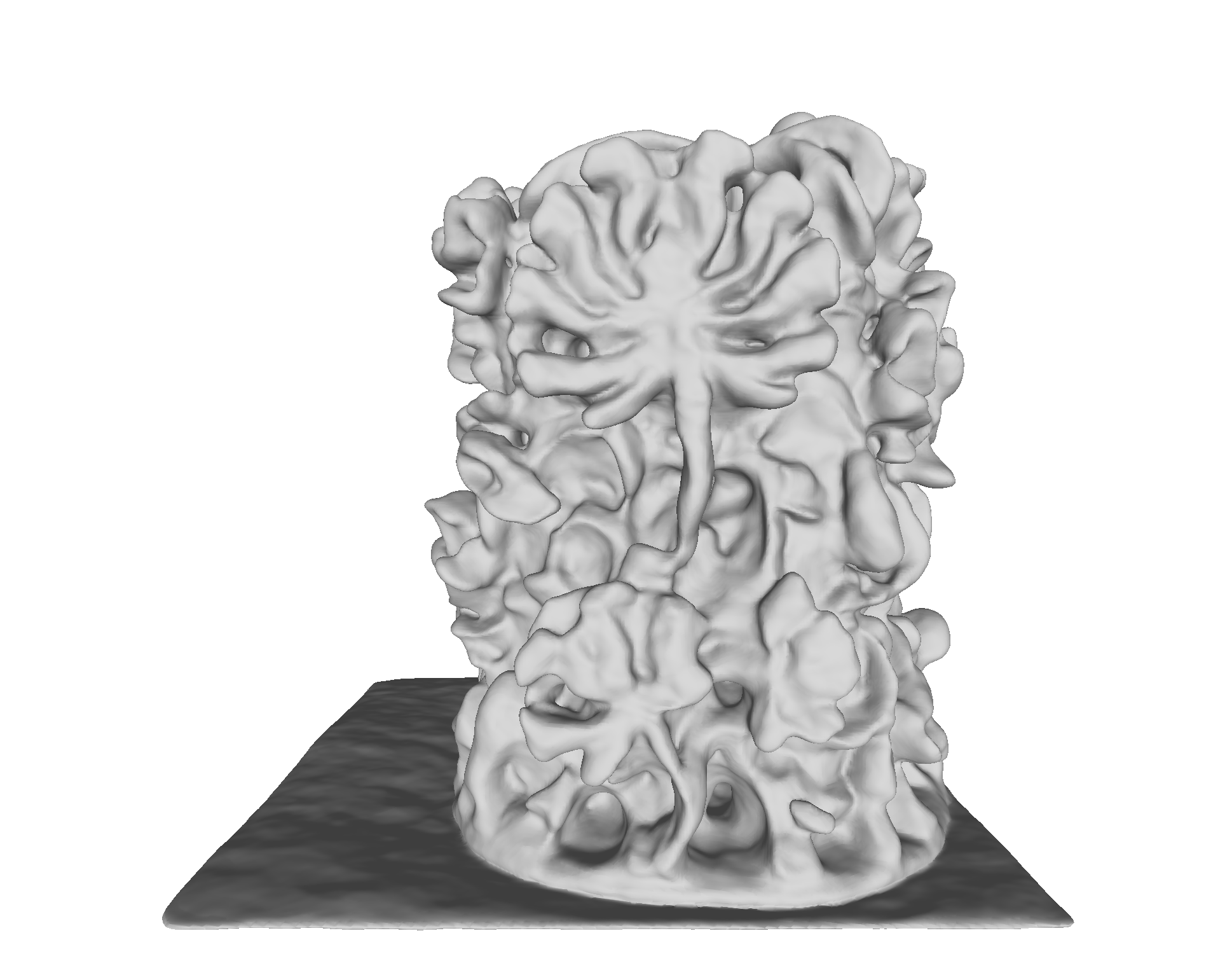}}
\end{minipage}
\begin{minipage}{0.23\linewidth}
    \vspace{3pt}
    \centerline{\includegraphics[width=\textwidth]{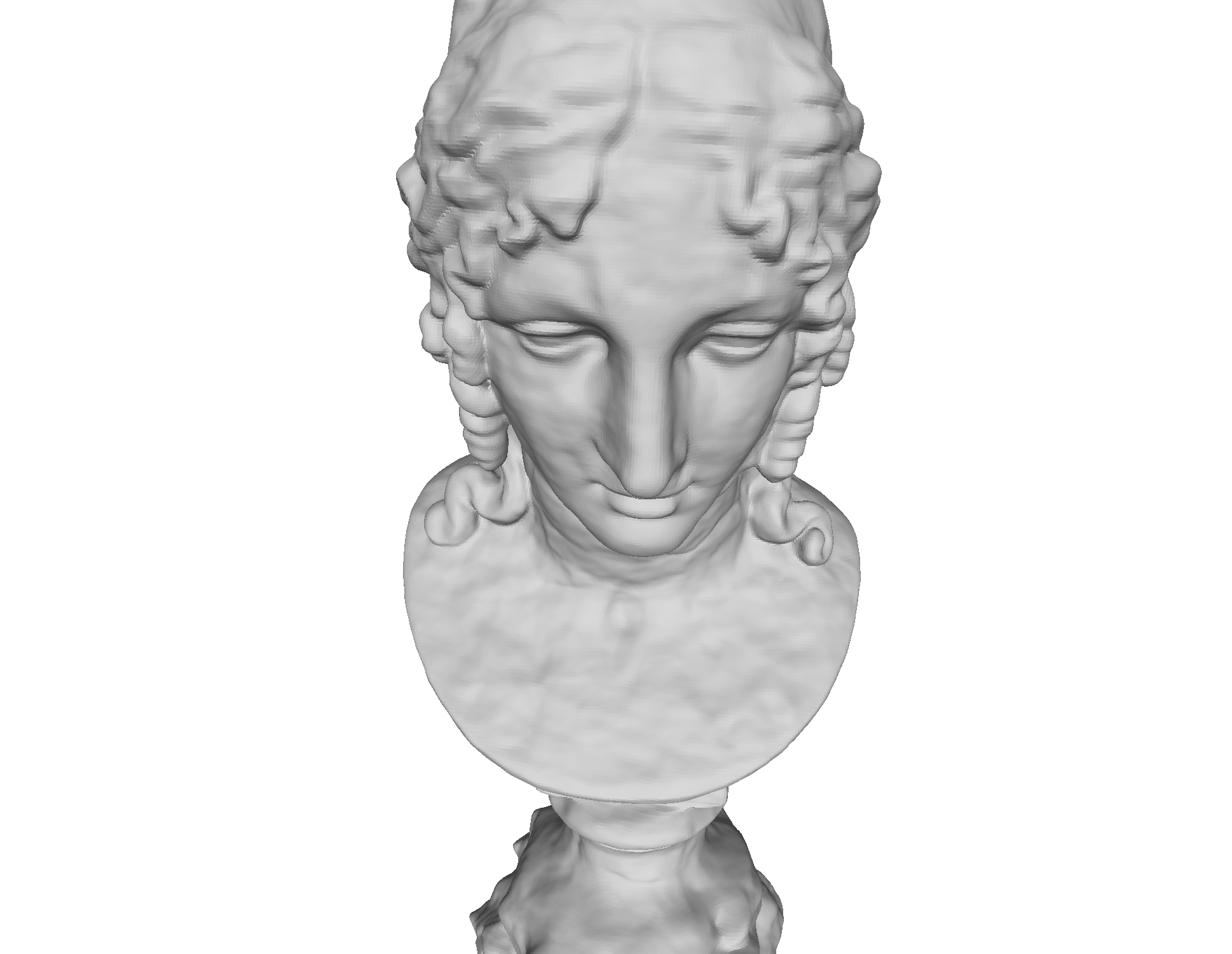}}
\end{minipage}
\begin{minipage}{0.23\linewidth}
    \vspace{3pt}
    \centerline{\includegraphics[width=\textwidth]{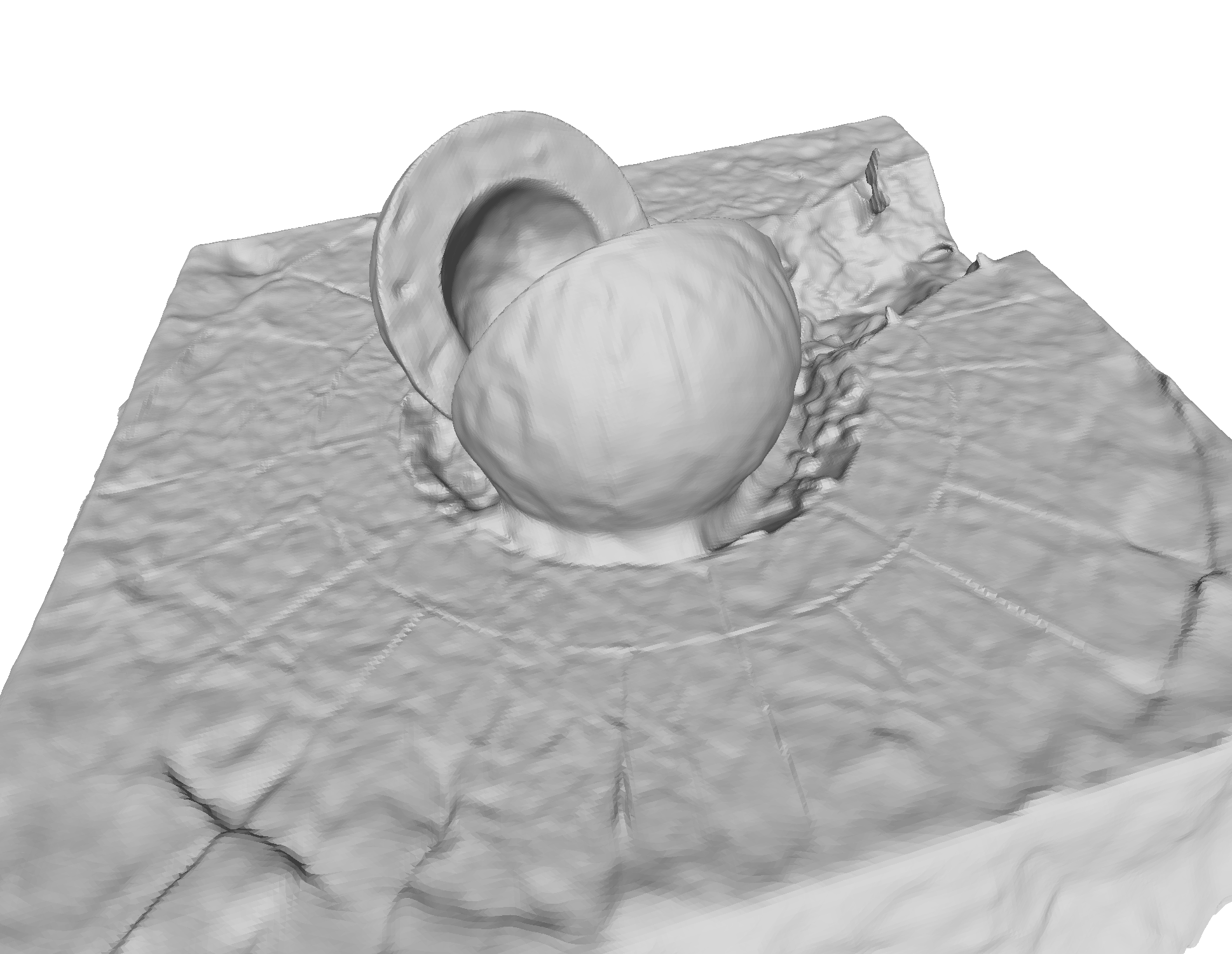}}
\end{minipage}

\caption{Qualitative comparison of NeuS* and BundleRecon on BlendedMVS}
\label{fig:qualitative_bmvs}
\end{figure*}

We trained NeuS+BundleRecon on the BlendedMVS dataset with masks. Moreover, the qualitative results are shown in Fig.~\ref{fig:qualitative_bmvs}. In the first scene, BundleRecon removes the geometric depressions on the dog's leg. In the second scene, we can see that BundleRecon fills the incomplete petals in the lower left corner of the jade. In the third scene, BundleRecon provides more refined hair textures. And in the last scene, BundleRecon better handles the concave part of the stone.

Table \ref{table:vs} summarizes the quantitative results of BundleRecon on the DTU dataset. IDR* and IDR+BundleRecon are trained with maskes, while NeuS* ans NeuS+BundleRecon are trained without masks. The table records both the chamfer distance of the reconstructed meshes and the PSNR of the rendered images, which clearly indicate that incorporating BundleRecon leads to improved performance for both IDR* and NeuS*. These results also demonstrate the compatibility of BundleRecon with the existing methods.


\begin{table*}[htbp]
\caption{Ablation Study on Loss Functions.}
\label{table:ablation}
\centering 
\begin{tabular}{c|c|c|c|c|c|c|c}
\hline
\textbf{Method} & \textbf{Bundle} & \textbf{M+V $l_1$} & \textbf{M+V $l_2$} & \textbf{Laplace} & \textbf{Sobel} & \textbf{normal} & \textbf{Chamfer} \\ \hline
exp1            & \Checkmark       &                    &                    &                  &                &                 & 2.07             \\
exp2            & \Checkmark       & \Checkmark          &                    &                  &                &                 & 2.08             \\
exp3            & \Checkmark       &                    & \Checkmark          &                  &                &                 & \textbf{1.98}    \\ \hline
exp4            & \Checkmark       &                    & \Checkmark          & \Checkmark        &                &                 & 2.15             \\
exp5            & \Checkmark       &                    & \Checkmark          &                  & \Checkmark      &                 & \textbf{1.82}    \\ \hline

\end{tabular}
\end{table*}

\begin{table}[htbp]
\caption{Results on Different Ray Bundle Settings.}
\label{table:bundleablation}
\centering
\begin{tabular}{c|c|c|c}
\hline
\textbf{Method} & \textbf{BundleSize} & \textbf{BundleNum} & \textbf{Chamfer} \\ \hline
exp6            & $ 3 \times 3$       & 229                & \textbf{1.82}    \\
exp7            & $ 5 \times 7$       & 229                & 1.92             \\
exp8            & $ 7 \times 7$       & 229                & 2.09             \\ \hline
exp9           & $ 3 \times 3$       & 57                 & 2.15             \\
exp10           & $ 3 \times 3$       & 114                & 1.94             \\
exp11           & $ 3 \times 3$       & 229                & \textbf{1.82}    \\ \hline
\end{tabular}
\end{table}

\subsection{Ablation Study}
\label{subsec:ablation}

Firstly, we invstigate the effectiveness of our loss functions through ablation studies. The first experiment involves replacing the single ray with the ray bundle while still utilizing the original loss function of IDR. In exp 2 and exp 3, we introduce the additional constraints for mean and variance, with different norms applied for these constraints. Furthermore, in exp 4 and exp 5, we utilize the different kernels to extract convolutional features. All these experiments are conducted on DTU scan 24. The results of our ablation study on the loss functions are presented in Table \ref{table:ablation}. It can be observed that the $l_2$ norm outperforms $l_1$ norm in terms of mean and variance constraints. More importantly, the Sobal kernel, which introduces the first order derivative information, performs better than the Laplace convolution kernel.

We also perform the ablation study on ray bundle settings with DTU scan 24, and present the results in Table \ref{table:bundleablation}. The experimental results demonstrate that enlarging the size of the ray bundle may result in poor performance with a fixed number of ray bundles. This is because the excessive information around the pixels can impede the model from converging effectively. Therefore, a larger size of ray bundle is unnecessarily better. However, we observed that increasing the number of ray bundles while keeping the bundle size fixed can lead to better reconstruction results. This indicates that, if the memory cost issue can be resolved, adopting more ray bundles could yield further improvements in reconstruction quality.

\section{Conclusion}
\label{sec:conclusion}

\begin{figure}[htbp]
    \centering
    \includegraphics[width=7cm]{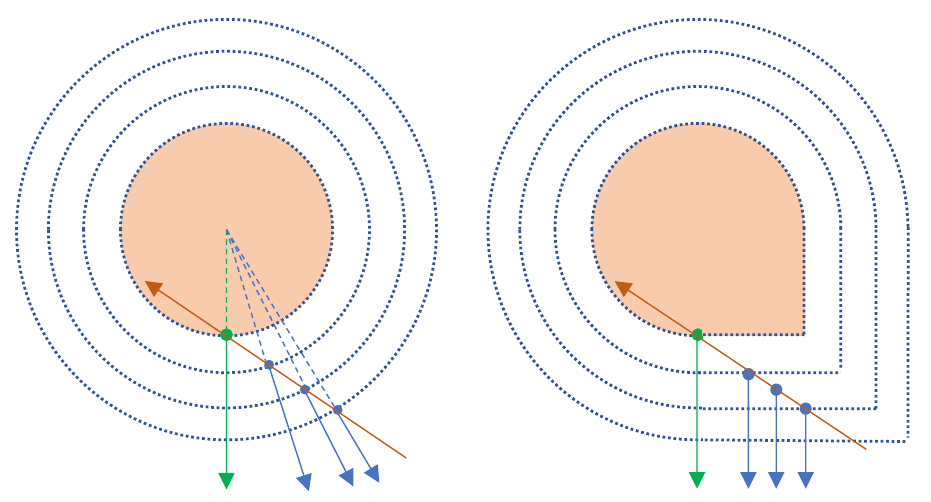}
    \caption{The green point represents the actual surface point, while the blue points represent the sampled points along the ray. As shown in the left figure, the points along the ray have different normals. If we artificially force these normal vectors to be the same, it will result in geometric distortion, as shown in the right figure. }
    \label{fig:normal}
\end{figure}

We present BundleRecon, a novel 3D reconstruction module for multiview implicit reconstruction that incorporates information from adjacent pixels. Our method is accompanied by a set of bundle-based loss functions to effectively constrain it. The experiments demonstrate that BundleRecon is compatible with the existing single ray based neural implicit models and can be seamlessly integrated into them to enhance their reconstruction quality.


Our future research will focus on reducing the memory cost associated with the ray bundle to match the number of single rays used in current works. This will further improve reconstruction quality. More importantly, we found that introducing ray bundles makes it possible to impose direct constraints on geometry network. Below we present a possible constraint approach, and there are still many details worth exploring.

As the training progresses, the sampled points gradually converge towards the true surface of the object. Based on the depth information and the ray direction, we can obtain the spatial coordinates of sampled points, which facilitates the normal computation. One possible way to impose constraints is to minimize the difference between the computed normal vector and the corresponding normal vector derived from SDF. However, this approach may cause geometric deformation. To illustrate this, we used 2D objects and their SDF isolines, as shown in Fig.~\ref{fig:normal}. When we sampled a series of points along the ray, these points had different normal vectors, as shown on the left. When we enforced the normal vectors of these points to be the same, the geometry was deformed, as shown on the right. We believe that a possible solution to this problem is to concentrate the sampling points near the surface and add normal vector constraints at a later stage of training.

\section*{Acknowledgments}
This work is supported by National Natural Science Foundation of China under Grants (61831015).

{\small
\bibliographystyle{ieee_fullname}
\bibliography{egbib}
}

\end{document}